# Process Monitoring Using Maximum Sequence Divergence

**Yihuang Kang · Vladimir Zadorozhny**

**Abstract** Process Monitoring involves tracking a system's behaviors, evaluating the current state of the system, and discovering interesting events that require immediate actions. In this paper, we consider monitoring temporal system state sequences to help detect the changes of dynamic systems, check the divergence of the system development, and evaluate the significance of the deviation. We begin with discussions of data reduction, symbolic data representation, and the anomaly detection in temporal discrete sequences. Time-series representation methods are also discussed and used in this paper to discretize raw data into sequences of system states. Markov Chains and stationary state distributions are continuously generated from temporal sequences to represent snapshots of the system dynamics in different time frames. We use generalized Jensen-Shannon Divergence as the measure to monitor changes of the stationary symbol probability distributions and evaluate the significance of system deviations. We prove that the proposed approach is able to detect deviations of the systems we monitor and assess the deviation significance in probabilistic manner.

**Keywords** Process Monitoring · Anomaly Detection · Sequence Data Mining · Process Mining · Information Theory

---

Yihuang Kang (✉)

Department of Information Management, National Sun Yat-sen University

70 Lienhai Rd., Kaohsiung 80424, Taiwan

E-mail: ykang@mis.nsysu.edu.tw

Vladimir Zadorozhny

School of Information Sciences, University of Pittsburgh

Pittsburgh, PA 15260, U.S.A.

E-mail: vladimir@sis.pitt.edu

## 1 Introduction

Various on-line information systems continuously create event logs that represent conditions of dynamic systems in different times. For example, Electrocardiography logs activity of the heart over a period of time; a disease outbreak detection system records the number of outpatient visits for some particular diseases; and a credit card fraud detection system monitors suspicious credit activities. These event logs from the information systems contain patterns of interest that can be identified by domain experts, discovered by pattern classification methods, or simply represented in meaningful symbols. By monitoring these dynamics of patterns across the times, we can understand the changes of a system and take required actions if necessary, which motivates us to develop *Process Monitoring* techniques.

We define the process as *a series of activities or state transitions of a dynamic system that produce some specific, either deterministic or probabilistic, outcomes*. Process monitoring refers to tracking development of system and evaluating the conformance of the development with expected or existing one. Many approaches related to the process monitoring are proposed in different fields of studies, such as *anomaly detection* [1], *change-point detection* [2]–[6] and *statistical process control* [7]. Anomaly detection focuses on finding "the patterns in data that do not conform to a well-defined notion of normal behavior" [1]; Change-point detection emphasizes the discovery of abrupt changes by comparing data in two adjacent time frames; and the statistical process control concerns the control of the processes by identifying the source of process/product variations. Here, however, we assume we monitor a dynamic system that keeps yielding the data that represents its states in different times. The sequences of these states indicate the development of the system and we are interested in finding the deviation of the system development from the discrete temporal sequences.

Due to the fact that the majority of the data generated by information systems are often numeric time-series, many existing approaches aim to directly cope with these series data streams. For example, the Shewhart Control Chart[3], Cumulative SUM (CUSUM)[3], and the Generalized Likelihood Ratio[2] focus on detecting shifting means, outliers, and high likelihood ratios to previous learned models. These approaches and their extensions have been actively discussed in data mining communities for years and are widely used in many real-world applications. However, the explosion of the data dimensions and numerosities in the recent years impedes the performance of these approaches, because the processing of high dimensionality/numerosities data requires more computational power as the amount of data grows. Many researchers have started considering the data dimensionality and numerosity reductions using pattern classification methods [8] and time series representation techniques. The goal of these methods is to discretize the continuous features, keep the signatures (e.g. distance measure) of original data in the transformed space, and adapt the data into patterns of interest denoted by meaningful symbols [9]. These adapted symbolic data streams (discrete temporal sequences) can be regarded as the development of the system we monitor. Consider a simple temperature classification. We discretize temperature in degree centigrade C, ( $15 < C, 15 \leq C < 25, 25 \leq C$ ) into (**C**old, **W**arm, and **H**ot). Two sequences like (**C**,**C**,**W**,**C**,**W**,**W**,**H**,**H**) and (**C**,**C**,**W**,**W**,**W**,**W**,**H**,**H**) reflect the transitions of the weather condition in terms of the feeling of the temperature during different monitoring periods. We consider these sequences of symbols as the states of the system and use these sequences to assess the deviation of the system development.

The similarities among these sequences in different monitoring periods provide us the information about the deviation of a system. The most intuitive way of measuring the differences among these sequences is to calculate the number of (mis)matched symbols in terms of symbol positions between two sequences. For example, the *longest common subsequence* [10] measures the similarity between sequences by finding the common subsequence, and the *edit distance* [11] evaluates the difference between sequences by counting the required operations to match two sequences. On the other hand, those distribution-based approaches, such as *Kullback-Leibler Divergence* ($D_{KL}$) [12], *Bhattacharyya distance*[13], and *Jensen-Shannon Divergence* ($D_{JS}$)[14], which measure the differences among the sequences by comparing discrete symbol probability distributions, can also be used as distance measures. However, some properties of these measures, such as *boundlessness* and *asymmetry*, make them inappropriate to be the criteria that help determine the system deviations. To track and assess the deviations, we suggest using a measure that is: 1) *Bounded*, which provides certain limits of the deviation that simplify the magnitude evaluation when the measure is used in numerical applications; 2) *Symmetric*, which ensures the identity of the deviation for the same set of probability distributions (i.e. different permutations of the same set of distributions have identical deviation); 3) *Generalizable*, which allows for the comparisons of multiple distributions/sequences in different times; 4) *Weightable*, which enables itself to assign different weights to different distributions for various applications (e.g. recent symbol probability distributions weight higher if we believe they are more important). In this paper, we use generalized Jensen-Shannon Divergence ($D_{GJS}$) [14] as the deviation measure used in the process monitoring, as $D_{GJS}$ possesses all the aforementioned properties.

To estimate the symbol probability distributions for the divergence/distance measures, the simplest way is to create a relative frequency vector (FV) by counting the frequencies of these symbols. However, a major downside of this approach is that it ignores the patterns of symbol transitions that may indicate the abnormalities of the system we monitor. Instead, we propose using stationary symbol probability distributions, which describe the probabilities of being at states (symbols) after the system have operated for a sufficient long period, from Markov Chains [15] generated by the discrete sequences. *Google's*

*PageRank* [16] algorithm is here used to obtain a unique stationary probability distribution for each sequence of symbols. These stationary symbol distributions capture the symbol transitions and represent the snapshots of the system dynamics, which also can maximize the aforementioned divergence measures used to assess the deviation of the system development.

The assessment of deviation significance is also crucial to the measures used in a monitoring system. The "assessment" here is simply the magnitude guideline of the measure we choose. It is to provide a critical threshold of when we should take the required actions (e.g. to give an alarm or make a decision). Many proposed approaches do not address this issue clearly. Some of them lack the threshold, set a predetermined threshold, provide a score function that computes a distance-like measure without the aforementioned properties, or suggest to choose a value that balances the quick detection and the false alarm rate [3], [5], [6]. In the worst case, the thresholds of these approaches need to be changed when these approaches are applied to different datasets, which hinders them from being applied to real-world applications. The reason we choose $D_{GJS}$ is that the approximation of its significant threshold can be obtained [17]. No matter what the symbolic data stream is, we can derive the statistically significant threshold for $D_{GJS}$ given a number of different symbols we have in the sequences, the number of symbol probability distributions we compare, and a significant level $\alpha$ (usually 0.05 or 0.01) we set. The significant level can also be interpreted as "the probability that the deviation ($D_{GJS}$) of the system development is higher than the threshold", which is valuable as a reference to decision makers. By monitoring the $D_{GJS}$, we can not only track the deviation but also assess the significance of the deviation of a dynamic system in probabilistic manner.

The contributions of this paper are summarized as follows:

— We introduce a novel approach used in process monitoring that helps detect the anomalies of a dynamic system from the point of views of both system change-point and long-term evolutionary deviation identified in discrete temporal sequences.

— We show that comparing stationary symbol probability distributions generated by Google's PageRank algorithm instead of the discrete symbol probability distributions from the frequency of symbols can maximize the information divergences.

— We present that generalized Jensen-Shannon Divergence outperforms other measures in terms of the accuracy of system change-point/deviation detections.

— We demonstrate that the significant threshold of the generalized Jensen-Shannon Divergence can be used as a criterion to determine the anomalies in sequences.

In addition, we also discuss the roles of four important properties (i.e. Boundedness, Symmetry, Generalizability, and Weightability) of a similarity/distance measure used in the assessment of system deviation.

The rest of this paper is organized as follows. In Section 2, we briefly review the backgrounds of existing approaches related to the process monitoring. The data reduction methods and the role of Markov Chain with the stationary symbol/state probability distribution are also discussed. In Section 3, we consider the proposed approach. In Section 4, we discuss the applicability and limitation of our approach by using synthetic and real-world datasets. We introduce other applicable distance measures commonly used in sequence anomaly detection and DNA/Protein sequence evolutions. In the end, we conclude with a brief summary of the advantages of using proposed approach.

## 2 Background and Related Work

The most intuitive related approach of process monitoring is to monitor one or more continuous random variables and see whether they deviate from the target/expected values. The Shewhart Control Chart and CUSUM are the typical methods [3]. The Control Chart is a method that keeps sampling the system and checking whether the sample means exceed a certain number of standard errors from the means. The CUSUM is similar to the Control Chart but it keeps calculating the cumulative sum of the differences to the target value until the sum exceeds a certain threshold. These distance-to-target approaches are easy to implement, but are unable to discover the lurking presumable patterns that may result in the significant variances before the system deviations actually occur. The likelihood ratio-based approaches [2], [6] are proposed to eliminate these shortcomings. These approaches provide the ratio statistics by comparing the fits of the models in different times. Then, the statistical tests on these ratios are performed or a scoring function is used to identify the significance of the deviations.

As discussed in Section 1, those approaches that directly handle the real-valued time series data are subject to the *Curse of Dimensionality* [8], [18] as the amount of data and the data dimensions have grown dramatically in recent years. Many data reduction methods are proposed to solve the problem. For example, Principal Component Analysis[19] (PCA), Discrete Fourier Transform [20], Singular Value Decomposition [21], Piecewise Aggregate Approximation (PAA) [22], Shape Definition Language(SDL) [23], and Symbolic Aggregate approXimation(SAX) [24] are proposed to reduce the data dimensionality and numerosity. Here we focus on those techniques that symbolize the raw data into sequences of symbols, as the benefits of analyzing the symbolic data stream are both the dimensionality /numerosity reduction and the measurement noise-insensitivity [9]. Also, numerous sophisticated sequence analysis methods, such as Permutation, Bernoulli, and Markov models [25], can be used to efficiently manipulate and perform the analysis on the symbolic data stream [26]. To preserve the essential information in the original data, the data reduction method we choose must also be able to keep the signatures (e.g. the distance measure) of the original data in transformed data space as discussed in Section 1. That is, if two series data are found similar in the original space, they should also be found similar in the transformed space. This property of a discretization method is known *lower bounding* [27]. In this paper, we demonstrate our approach by assuming that we deal with a dynamic system that generates continuous time-series data stream. We choose simple cut-points and the SAX as examples to perform the data discretizations.

SAX is a method that discretizes a univarate real-valued time series and produces symbols with approximately equal probabilities. The time-series data is divided into *i* segments of equal length. Given a normalized time series data, the distribution is divided into *k* equal-probable areas that are assigned *k* possible symbols. Each equal-length segment in the data is replaced with a symbol based on which area the average value of the segment is in. Figure 1 shows an example of how SAX works.

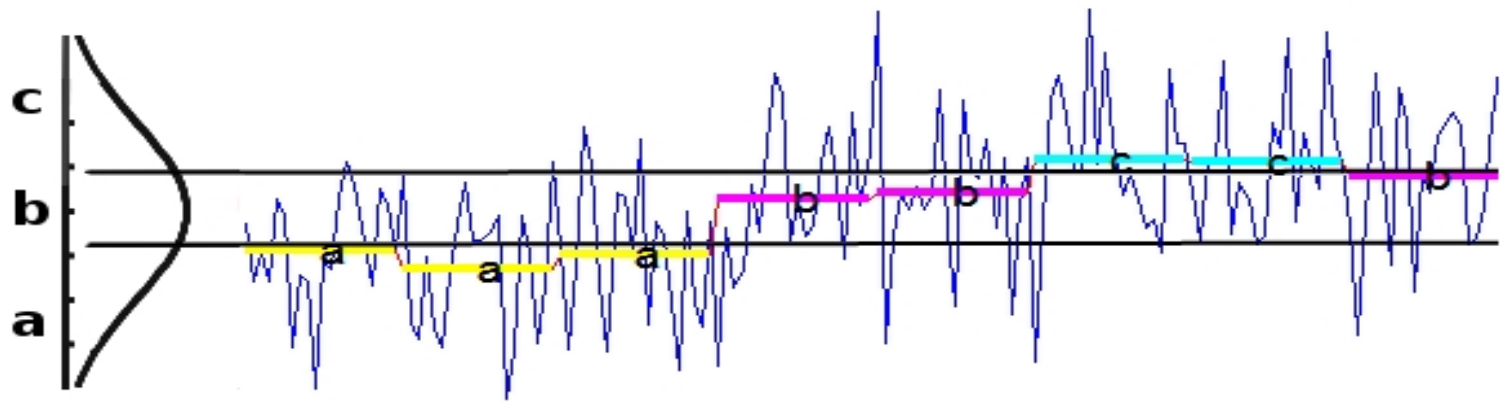

**Figure 1: Data discretization by SAX**

In Figure 1, the series is divided into 8 segments of equal length. The distribution of the series is divided into 3 equal-probable areas that are represented by 3 letters (***a***, ***b***, ***c***). Each segment is assigned a letter/symbol (***a***, ***b***, or ***c***) based on where its average is located. The series data in Figure 1 is then mapped into ***aaabbccb***. One of important applications of SAX is to discover the time series discords [24], which is to find the unusual patterns/subsequences within a time series. We use SAX in our experiments because it has the abovementioned advantages—numerosity reduction and lower bounding [24], [28]. Unlike this particular application of SAX to find discords, our approach uses the sequence of symbols generated by SAX to detect the significant changes and long-term deviation of the symbol probability distributions. That is, we are interested in finding the deviation of a dynamic system, not the specific abnormal patterns/sequences.

We consider building probabilistic process models and estimating the probability distributions for the symbolic data streams generated from the data reduction/discretization methods. Similar to the notion of *Conformance Checking* in *Process Mining* [29], our approach also keeps constructing process models and finding the discrepancies between modeled behaviors and the newly observed ones. However, the naïve approach of Conformance Checking, Token Replay [29], focuses on the *fitness* of the models, which is the degree of whether the observed models can "replay" the deterministic process flows of the expected models. Instead, our approach aims at evaluating the significance of the system deviation in probabilistic manner by comparing the discrete symbol/state probability distributions in different times. Various existing probabilistic models can be used to build the process models and estimate the distributions. For example, the *Dynamic Bayesian Network* and the *Markov Networks* of the *Probabilistic Graphical Models* [30] construct the models based on the conditional dependent or independent structures among random variables. As we only tackle a discrete pattern/symbolic data stream, a discrete random variable that represents the states of a dynamic system, we consider a simple probabilistic model—the first-order Markov Chain.

The first-order Markov Chain we refer to is a discrete-time random process with the Markov property, which assumes that the next state of the system only depends on the current state [31]. The Markov Chain is a probabilistic model used to represent a discrete state-space dynamic system and is applied to various fields of studies. Here, we suggest creating Markov Chains from the symbolic data stream and estimating the symbol distributions by obtaining the stationary distributions. Different from some literatures doing the sequence analysis that estimates the symbol and symbol transition distributions by counting symbols' frequencies [17], [32], we use the stationary symbol distributions instead and assume the distributions can represent the snapshots of a system dynamics in different times. The benefit of doing this is that the abnormal system transitions can be captured and the evolutionary deviation of the system development can be discovered regardless of the actual number of occurrences of symbols and transitions. However, as some state/symbol transitions may not occur in any given time frame/sequence, which means not all the Markov Chain we created are ergodic, we may not be able to obtain a unique stationary symbol distribution for each chain [31]. To solve this problem, we convert all the stochastic state transition matrices of the Markov Chains into *Google Matrices* [16], which are fully connected ergodic stochastic matrices and guaranteed to obtain the stationary distributions. The Google Matrix was originally used by *Google's PageRank* [33] algorithm to deal with very large sparse matrices that represent the links among web pages. We use the PageRank vectors as the stationary symbol distributions, since we are only interested in the symbol distributions, not the ranking of the symbol probabilities.

The generalized version of Jensen-Shannon Divergence ($D_{GJS}$) is used in our approach because of its valuable properties as discussed in Section 1. In recent decades, the $D_{GJS}$ is getting more popular and is extensively used as a divergence measure to the sequence analysis in many fields, such as the comparison of DNA sequence segments [32] in Bioinformatics and the distinguishability in quantum entanglement in Physics [34], [35]. Many extensions have also been developed to enhance the applications of $D_{GJS}$. For example, a Markovian form of the $D_{GJS}$ (MJSD) is proposed to deal with the sequences generated by Markov sources of arbitrary orders [32]. That is, the MJSD take the $n$th-order of symbol transitions ($n$th-order Markov Chain) into consideration, which is useful when higher order symbol transitions matter.

Unlike these applications of $D_{GJS}$, the proposed use of $D_{GJS}$ focuses on detecting any system transitions that may result in significant deviation in any given periods of system developments.

## 3 Proposed approach—Maximum Sequence Divergence

In this section, we consider proposed approach by beginning with the problem definition and discussion of symbol probability estimation. We assume that the monitoring system keeps receiving a numeric data stream from a dynamic system and symbolizing the data into sequences of symbols using data discretization methods.

### 3.1 Problem Definition and Symbol Probability Estimation

Consider, for example, that we have a symbolic data stream from a stock market index consisting of two possible symbols of changes ($k$ = 2), **U** and **D** as the index goes "**U**p" and "**D**own" for simplicity. We divide them into 5 equally-sized sequences of symbols as shown in Figure 2. These two symbols, **U** and **D**, are equal-probable in terms of relative frequencies in each sequence. At the very beginning ($S_1$), we can see that the index keeps iteratively up and down through the observation cycle. Then, the index becomes more stable. Our goal is to detect *the changes in discrete (symbol) temporal sequences*—the deviation of a system development. To obtain the stationary symbol probability distributions, we first build the first-order discrete-time Markov Chain with a symbol transition probability matrix (denoted by $\boldsymbol{H}_{\mathrm{m}}$) for each sequence by counting the relative frequencies of symbol transitions. We assume the development of this system is a random process of Markov property.

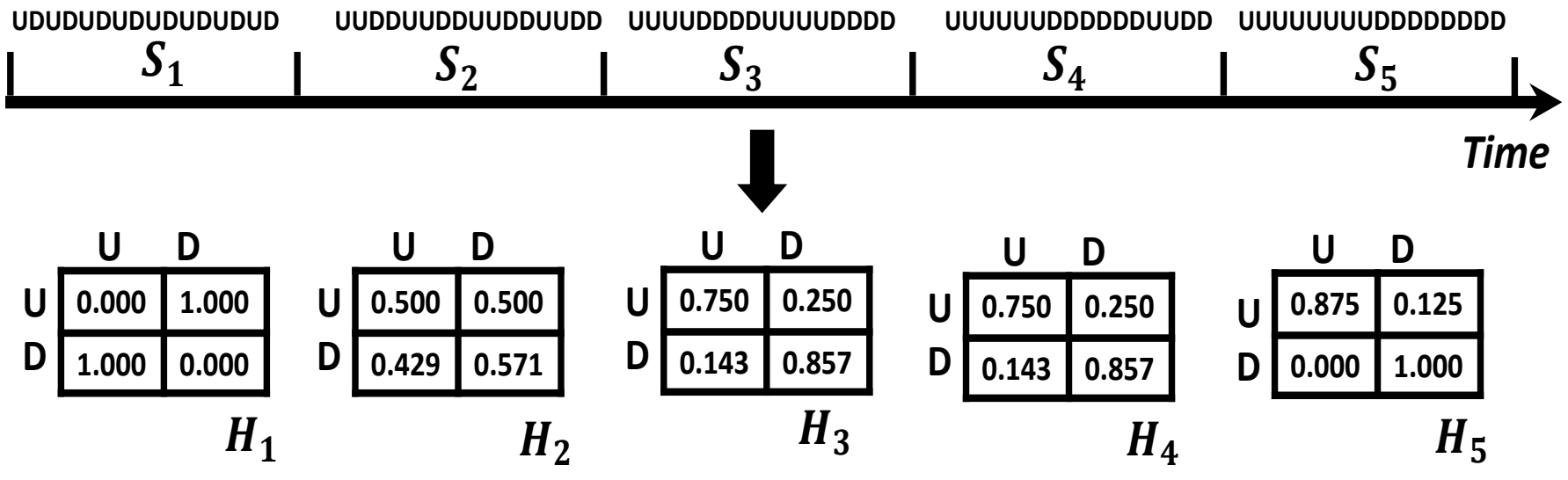

**Figure 2: Sequences and symbol transition probability matrices**

For each stochastic matrix of a Markov Chain, we can obtain a unique stationary probability distribution, also called *Steady-State Vector* (SV), if the matrix is ergodic [31]. That is, any state (symbol) can return to itself in one step and also can be reached from any other states in the stochastic matrix. In this case, the Markov chain we created is fully connected and all transitions have a non-zero probability.

Evidently, there is no guarantee that we can create such matrices from all the sequences, as some state transitions may never occur within a time frame (sequence). To solve this problem, we propose constructing the stochastic matrix for each sequence and then converting these matrices into Google Matrices [16]. The Google Matrix is an ergodic and stochastic matrix originally used by Google's PageRank algorithm to deal with very large sparse matrices that represent the links between web pages. The Google Matrix **G** can be computed as:

$$\boldsymbol{G} = d\boldsymbol{H} + (d\boldsymbol{a} + (1-d)\boldsymbol{e})\frac{1}{k}\boldsymbol{e}^T \quad \textbf{(1)}$$

where $\boldsymbol{H}$ is the original stochastic (symbol transition) matrix created from a sequence. The **a**, **e**, and $k$ denote the binary dangling node vector, the rank-one teleportation vector, and the number of possible states/symbols respectively, and $d$ is the damping factor that is between 0 and 1. Note that we use $d$ instead of $\alpha$ found in most literature in order to be distinguished from the statistical significant level $\alpha$ used in the following sections. As the $\boldsymbol{G}$ is dense and fully connected, we can then obtain a unique SV, also called the *PageRank vector* [16]. Instead of being used to rank the pages, the generated PageRank vectors are considered the symbol probability distributions in our approach.

The SVs contain fixed probability of each state (symbol) when a Markov chain operates for a sufficiently long period [31]. Here, the continuously-created stochastic matrices and the SVs can be considered snapshots of the system transitions and evolutions for long run. One advantage of considering the changes of the symbol probability distributions in SVs instead of those from the frequency of symbols in the sequences is that the SVs also take the orders of symbols (transitions) into consideration, which is valuable when SVs are used in the detection of abnormal transitions. In Figure 3, we show how to create the SVs from these 5 sequences ($S_1, S_2, S_3, S_4, S_5$) shown in Figure 2. We convert $\boldsymbol{H}$ into $\boldsymbol{G}$ with damping factor $d$ = 0.99. For each $\boldsymbol{G}$, we can obtain a unique SV. These 5 sequences are then transformed into a series of 5 SVs.

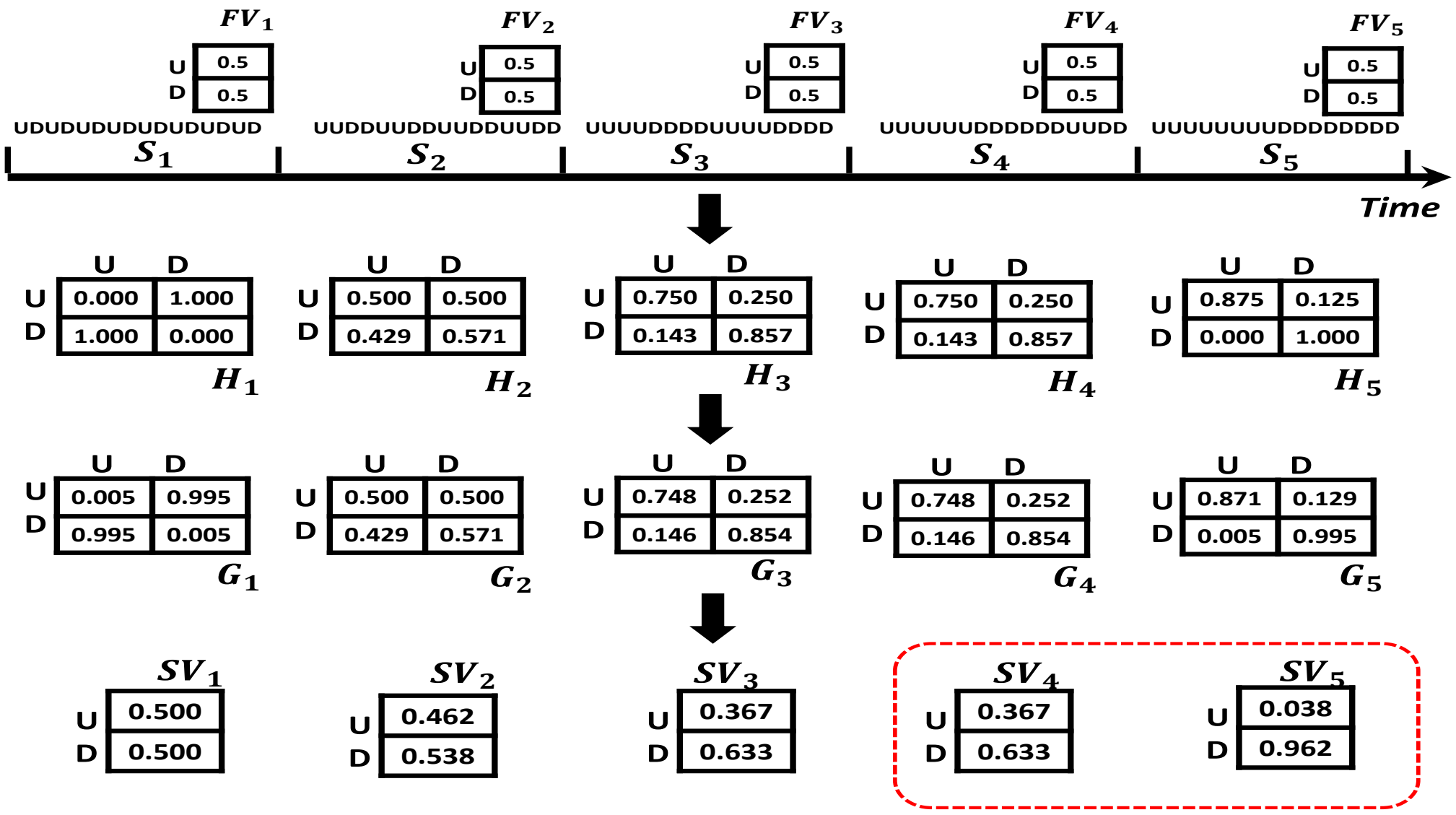

**Figure 3: Conversion from sequences to steady-state vectors**

Note that the damping factor ($d$ = 0.99) plays an important role here. It is originally used to control the rate that the random page surfers follow the hyperlink structures or jump to a random new page [16]. Here, we consider the damping factor the rate that adds/pads the probabilities of those lower-probable or zero-probable (absent) symbols transitions. The original damping factor used in the PageRank algorithm is 0.85, which balances the efficiency and the effectiveness of performing the power method to obtain the SV [16]. However, the choice of damping factor in our approach depends on how well the sequences we analyze reflect the actual dynamics of a system in different monitoring periods. That is, we use a lower damping factor if we believe those low- or zero- probable cells of a stochastic matrix should be higher, because the sequence we use may not represent the actual transitions of the system conditions. In most cases, we suggest using a high damping factor instead to avoid padding too high probabilities into these low- or zero-probable cells of a stochastic matrix so that we can maximize the differences/divergences among these SVs generated from different sequences, as a higher damping factor increases the sensitivity of the resulting vectors that are able to detect the smaller changes of the system[16]. In this paper, we use $d$ = 0.99 for all the experiments. Also, as the size of the stochastic matrix in our approach is determined by the number of possible states/symbols and is usually much smaller than the page link matrix (e.g. a 2 by 2 stochastic matrix in Figure 3), the high damping factor with small matrix does not requires significant computation time to obtain the steady-state vectors.

From Figure 3, we can see that the SVs change in terms of the probabilities of symbols (**U** and **D)** and show a trend that the market index is likely to go down at the end of the monitoring period. The approach to use the PageRank vectors, instead of relative frequency vectors (FV), captures possible different long-term symbol transitions and maximizes the deviation of system development in different time frames. Figure 3 also shows that there is not only a gradual deviation, but also a noticeable change between $SV_4$ and $SV_5$ (in dashed circle). The goal of the process monitoring is to detect both of them.

### 3.2 Measure Selection and Generalized Jensen-Shannon Divergence

By monitoring *Information Divergence* among the discrete probability distributions of these SVs, we are able to assess the deviation of the system. The "Information Divergence" here is the notion of distance that indicates the difference among two or more probability distributions. Most divergence measures do not satisfy the strict conditions as a true distance metric in mathematics, i.e. the *symmetry* and *triangle inequality*, which means these divergences should not be used as a regular distance metric to compare arithmetically. To select an appropriate measure, we define the first two requirements of a distance/divergence $D(P_x,P_y)$ we need:

$$D(P_x, P_y) \geq 0 \quad \textbf{(2)}$$

$$D(P_x, P_y) = 0, \text{iif } P_x = P_y \quad \textbf{(3)}$$

where $P$ is a discrete probability distribution (i.e. the SV in our approach). P = $[p_1, p_2, ..., p_k]$ and $\sum_k p_k = 1$. At first glance, we can just use a divergence that meets the Eq. 2 and Eq. 3 in our monitoring system. However, as discussed, there are some popular divergences widely used in various fields, but not all of them are appropriate deviation measures we need. Take the Kullback-Leibler Divergence ($D_{KL}$), also known as *relative entropy*, as an example. The divergence is defined as:

$$D_{KL}(P_x, P_y) = \sum_{\mathrm{i}=1}^{k} P_x(\mathrm{i}) \log_{\mathrm{k}} \frac{P_x(\mathrm{i})}{P_y(\mathrm{i})} \quad \textbf{(4)}$$

where the base $k$ is the number of discrete probabilities (the number of components in an SV). If we have two SVs, $P_1$=[0.5 0.5] and $P_2$=[0.9 0.1], for example, the $D_{KL}(P_1, P_2)$ = 0.737. It seems like $D_{KL}$ is the measure we need, but an asymmetric divergence like $D_{KL}$ cannot provide a common metric to evaluate the same set but different permutation of SVs. $D_{KL}$ is proved to be asymmetric and semi-bounded[14], which means:

$$D_{KL}(P_x, P_y) \neq D_{KL}(P_y, P_x) \quad \textbf{(5)}$$

$$0 \leq D_{KL}(P_x, P_y) \leq +\infty \quad \textbf{(6)}$$

For the previous example with two SVs, the $D_{KL}(P_1, P_2)$ is 0.737, but $D_{KL}(P_2, P_1)$ is 0.531. Also, there is no maximum limit of $D_{KL}$ for any given two probability distributions. These two properties (i.e. Eq.5 and Eq. 6) make $D_{KL}$ an inappropriate measure. Again, our goal is to assess the significance of divergence/deviation for a system by monitoring a measure from these continuously-created SVs. Here, we restate the two required properties of the divergence. A divergence $D(P_x, P_y)$ we need must be *Bounded* and *Symmetric*, which means it must not only satisfy Eq. 2 but also Eq. 7 and Eq. 8 as follows.

$$0 \leq D(P_x, P_y) \leq a, a \in Q^+ \quad \textbf{(7)}$$

$$D(P_x, P_y) = D(P_y, P_x) \quad \textbf{(8)}$$

A bounded divergence provides certain limits of the deviation that simplify the magnitude evaluation when it is used in numerical applications, whereas a symmetric divergence ensures the identity of the deviation for the same set of symbol probability distributions, which means different permutations of the same set of probability distributions must have an identical deviation value—a generalized definition of the symmetry property. Note that the measure we need must also be able to cope with a set of probability distributions, as it is used in an online monitoring system to track the evolutionary deviation of a dynamic system. This requirement calls for the need of the other two important properties of a divergence—*Generalizability* and *Weightability* [36] as discussed below.

Consider another example, such as we have a real-time process monitoring system that keeps converting a symbolic data stream into *m* SVs $(SV_1, SV_2, \ldots SV_m)$ with 3 different states/symbols ($k$ = 3) as shown in Figure 4. We need a measure that can assess the deviation of the system by continuously calculating the divergences from the changes of discrete probability distributions for a set of SVs. That is, the divergence measure must be able to be generalized to compare multiple discrete probability distributions in different times—the generalizability property of a measure. In Figure 4 , for example, the divergence should allow for the comparisons of $D(SV_1, SV_2)$, $D(SV_1, SV_2, SV_3)$, $D(SV_1, SV_2, \ldots SV_m)$, i.e. any combination of the SVs. Also, a practical application of online monitoring systems is that only part of (usually the most recent) developments/activities of a system is important. That is, more recent activities of a system weigh higher. We suggest that a divergence measure should also be able to assign a weight value $\boldsymbol{\pi}_m$ for each distribution we compare—the weightability property of a measure. For example, if we believe that monitoring the divergences that compare the latest 4 SVs in Figure 4 is enough to evaluate the system deviation, we can only assign the weights ($\boldsymbol{\pi}$) to these 4 SVs and keep the weights of the rest SVs as 0.

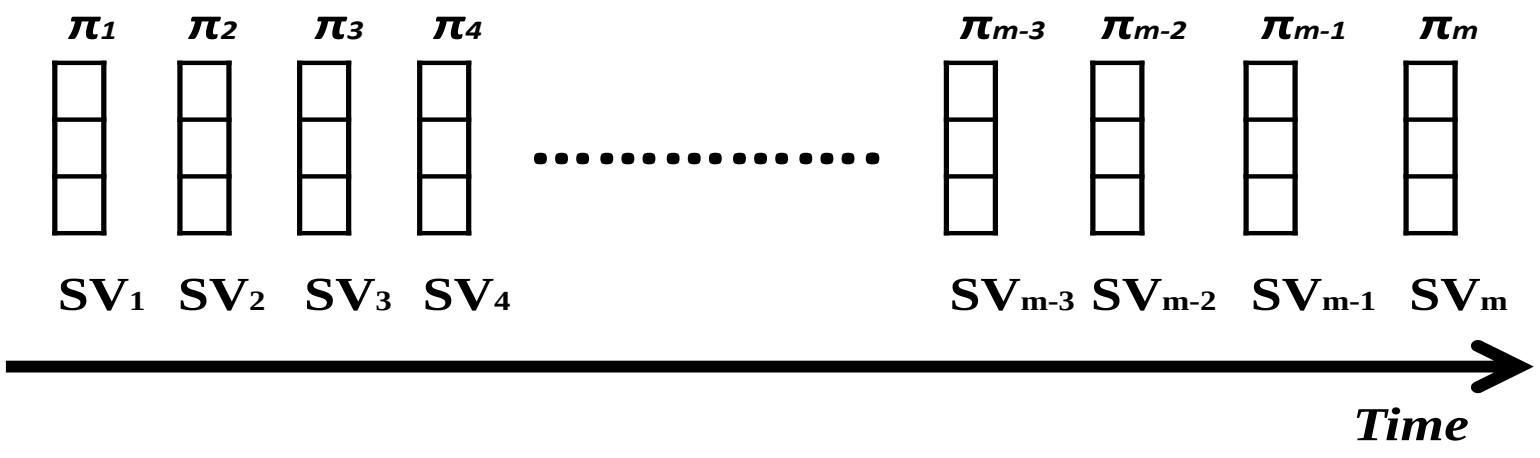

**Figure 4: A series of steady-state vectors with weights**

We chose generalized Jensen-Shannon Divergence ($D_{GJS}$) [14] as the measure used in our process monitoring approach, because the $D_{GJS}$ possesses all of the four properties mentioned above. $D_{GJS}$ is a symmetric measure that ranges between 0 and 1 [14], [35]. The $D_{GJS}$ is defined:

$$D_{GJS}(P_1, P_2, \dots, P_m) = H\left(\sum_{i=1}^{m} \pi_i P_i\right) - \sum_{i=1}^{m} \pi_i H(P_i) \qquad \textbf{(9)}$$

where $\pi_i$ is the weight and $\sum \pi_i = 1$. The $P_m$ are the discrete symbol probability distributions we compare. In our approach, $P_m$ are the SVs from sequences in different time frames. *H(x)* is the *k-ary Shannon Entropy* that is defined as:

$$H(x) = -\sum_{i=1}^{k} P(x_i) \log_k P(x_i) \qquad \textbf{(10)}$$

The $D_{GJS}$ can compare any finite number of the SVs. All the SVs can also be weighted. Take the five SVs in Figure 3 as an example, if we want to track gradual changes of the market ups and downs, we will take all the SVs into consideration. That is, we compute $D_{GJS}(SV_1, \dots, SV_5)$ with the equally-weighted $\pi = 0.2$ for all SVs. In this case, the $D_{GJS}$ is 0.1061. On the other hand, if we want to capture the recent abrupt changes of the market and believe that only the last two SVs (i.e. $SV_4$ and $SV_5$) are important, we compute $D_{GJS}(SV_4, SV_5) = 0.1362$ with $\pi_4 = \pi_5 = 0.5$. It is obviously that one can assign arbitrary weights to the SVs and then calculate $D_{GJS}$ from any combinations of SVs with different weights. However, literatures have indicated that the natural choice of the weights $\pi_i = (n_i / N)$, where $n_i$ is the length of the sequence used to estimate $P_i$ and $N$ is the length of total sequences used to compute $D_{GJS}$, may produce an optimal estimator of $D_{GJS}$ [14], [17]. Therefore, we suggest calculating selected SVs from equal-length sequences and assigning equal weights $\pi_1 = \pi_2 = \dots = \pi_m = (1 / m) = (n_i / N)$ to avoid the bias when estimating $D_{GJS}$. Besides, it is certain that higher $D_{GJS}$ indicates higher deviation. However, we also need a magnitude guideline (i.e. how high $D_{GJS}$ is too high) for $D_{GJS}$ to assess the significance of the deviation so that we can make a decision based on it.

### 3.3 Deviation Significant Threshold and Assessment

There is no fixed value of $D_{GJS}$ as a threshold that indicates the divergence is "high enough" to take action. From the definition of $D_{GJS}$, we can see that the $D_{GJS}$ varies dramatically based on 3 factors—(i) the number of components in an SV $k$ (the number of different symbols); (ii) the number of distributions/SVs $m$ we compare; (iii) the weights for all the distributions/SV $\boldsymbol{\pi}$. That is, even for the same symbolic data, the number of different symbols/states $k$ we choose when we symbolize the raw data, the number of sequences/SV $m$ we compare, and the weights of SVs $\boldsymbol{\pi}$ we assign, can noticeably increase or decrease the $D_{GJS}$. Here, we first illustrate how the number of states $k$ affects the $D_{GJS}$ by an example. Suppose we have two SVs, $SV_1$ and $SV_2$. Both of them have $k$ probabilities (for $k$ states/symbols) and a dominant state with the probability of 1.0. But, the dominant state in $SV_1$ is the first state, whereas the dominate state in $SV_2$ is the second state. The probabilities for the rest of the states for $SV_1$ and $SV_2$ are all zero. These two SVs are equally-weighted, i.e. $\pi_1 = \pi_2 = 0.5$. Figure 5 shows the $D_{GJS}$ decreases as $k$ increases when we compare these two distributions.

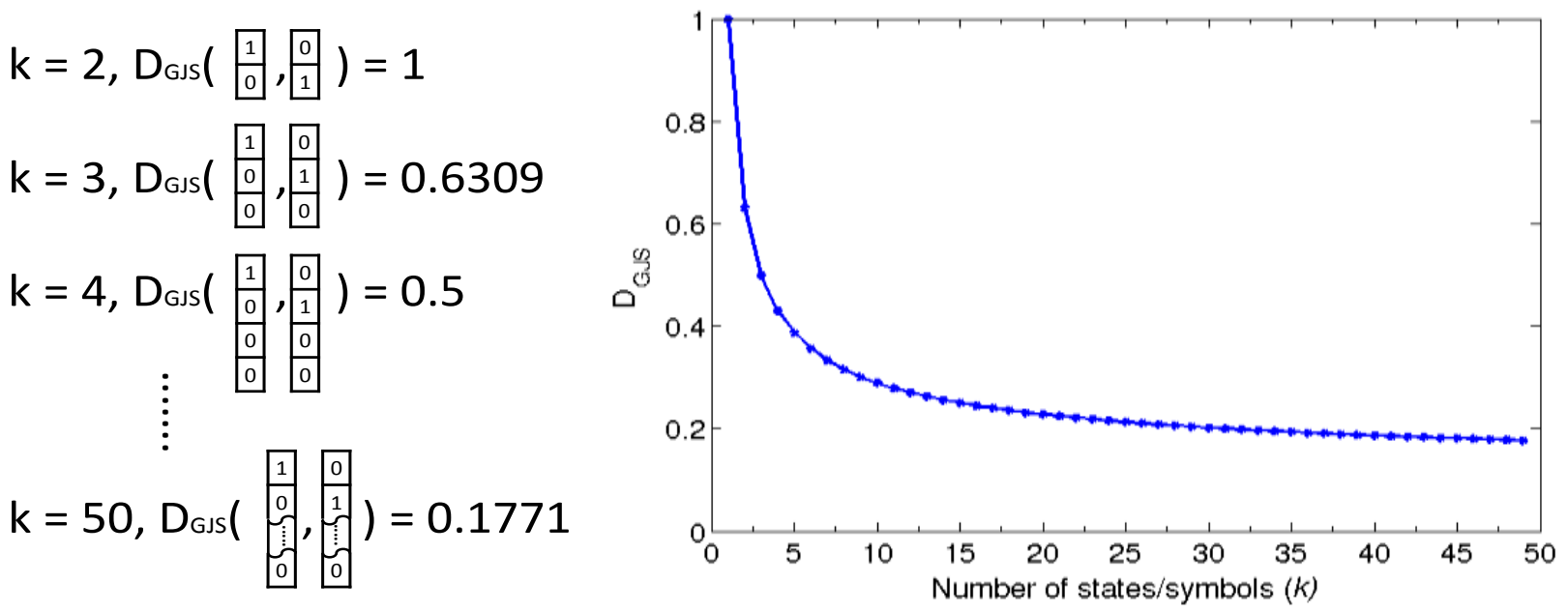

**Figure 5: $D_{GJS}$ vs. number of different states ($k$)**

Apparently, we can explain the negative association by the change of the data granularity. When applying a data discretization method to symbolize the data, a higher number of different symbols (the number of states in an SV) will increase the granularity and proportionally diminish the differences among the symbol probability distributions we compare. We can also explain it by the definitions of the $k$-ary Shannon Entropy (Eq. 10). As the base $k$ increases, the entropy decreases. Correspondingly, the $D_{GJS}$ decreases as $k$ increases.

The number of SVs $m$ and the weights for these SVs $\boldsymbol{\pi}$ also have a great impact on the $D_{GJS}$. From the definition of $D_{GJS}$, we can see $D_{GJS}$ allows multiple weighted SVs (distributions). Consider a simple example that we have a monitoring system continuously comparing equally-weighted (i.e. $\boldsymbol{\pi}_1 = \boldsymbol{\pi}_2 = \ldots = \boldsymbol{\pi}_m = 1/m$) $m$ SVs created from a symbolic data stream. Figure 6 shows the example with the distributions of the last SV different from all previous SVs with $k = 3$. Note that the probability in second cell is always 1 instead of 0 in the last SV for different $m$ up to 50.

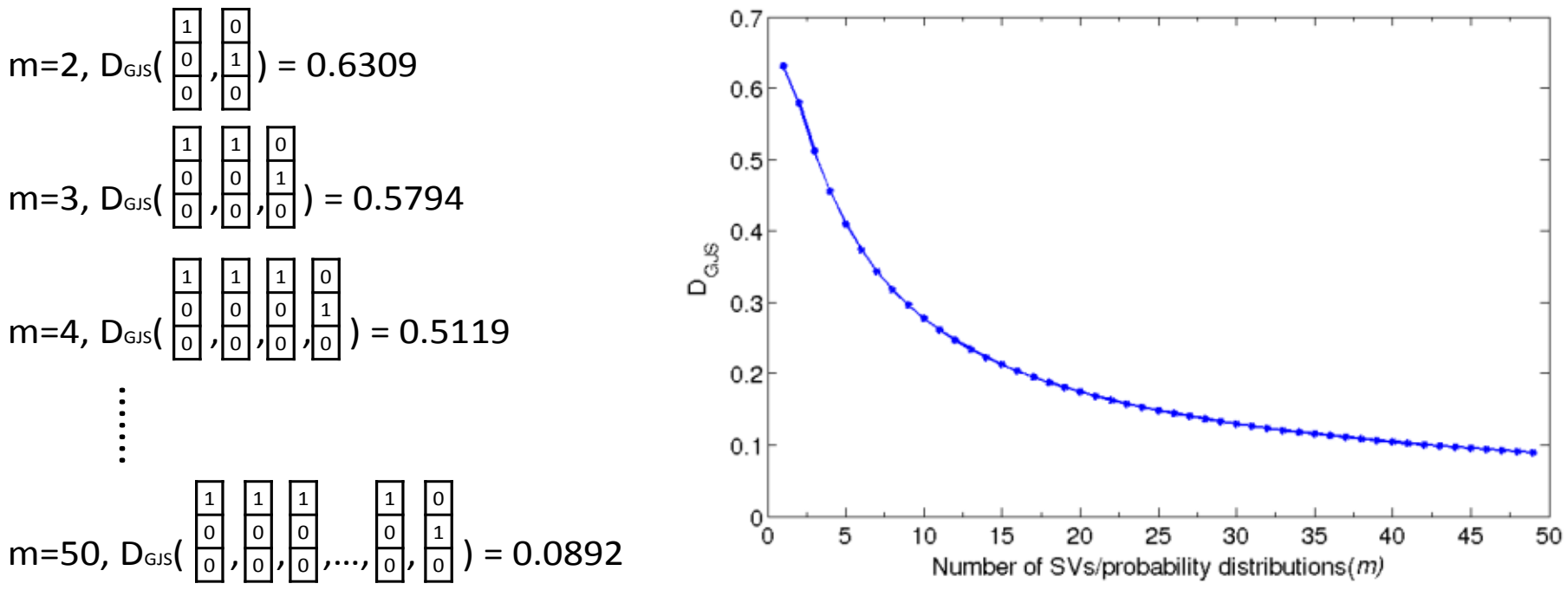

**Figure 6: $D_{GJS}$ vs. number of discrete probability distributions ($m$)**

We expect the monitoring system should report that the $D_{GJS}$ is "significantly high" for $m$ SVs when it compares the last SV with all previous ones, because the probability distribution of the last SV is noticeably different from previous SVs. However, it is clear the threshold to define the "significance" should also depend on $m$ and $\boldsymbol{\pi}$. Again, we can explain this by the definitions, i.e. Eq. 9 and Eq. 10. In the example, the weights $\boldsymbol{\pi}$ for all $m$ SVs are equal. If the number of SVs $m$ increases, the influence of each SV reduces so that the $D_{GJS}$ decreases. On the other hand, if we assign a very high weight to the last SV, the $D_{GJS}$ will increase dramatically as the influence of the last SV increases. Therefore, the number of the SVs $m$ and the choice of the weight $\boldsymbol{\pi}$ also play an important role when we determine the significant threshold of the $D_{GJS}$.

The significant threshold of the $D_{GJS}$ is a certain value of $D_{GJS}$ that answers the question—"what is the probability that the $D_{GJS}$ is higher than the threshold?". The probability here is the critical p-value (significant level $\alpha$) commonly used in Statistics. Here, we use $D_{GJS|k,m}$ to denote the threshold. Obviously, $D_{GJS|k,m}$ is essential for practical uses of $D_{GJS}$ as a deviation measure. Before introducing $D_{GJS|k,m}$, we first state our settings and assumptions again. The monitoring system continuously receives a symbolic data stream and divides it into sequences $S_m$ of total $N$ symbols with $k$ different possible symbols denoted by $A = (a_1, a_2, ..., a_k)$. The sequences $S_m$ are equally-sized $(S_1, S_2, ..., S_m)$ (i.e. the length of each sequence $n_1 = n_2 = ... = n_m$). We can then create $m$ first-order Markov Chains and the transition probability matrices from these sequences. These transition probability matrices are transformed into the Google Matrices $G_m$. As $G_m$ are ergodic and small, $m$ unique SVs can be easily obtained by the Power Iteration [16]. Then, we assign a weight $\pi$ for each SV depending on different applications as shown in Figure 4 to create a $k$ cells by $m$ SVs table ($k = 3$ in Figure 4). These SVs are the snapshots of the system we monitor and are of probabilities of these states (symbols). Consider this $k$ by $m$ table, we would like to know how much

*Information* that $k$ symbols and $m$ sequences/SVs share from this table—*Mutual Information* ($I$). In [17], the task of obtaining the $D_{GJS}$ is interpreted as the task of obtaining the Mutual Information in a symbol $a_k$ about an sequence $S_m$. That is, provided we know the probability distributions (SVs) of these $k$ symbols and what symbol $a_k$ we have drawn from these sequences, how much $I$ about which sequences $S_m$ we draw. In our approach, the Mutual Information $I$ is defined:

$$I \equiv D_{GJS}(P_1, P_2, \dots, P_m) \equiv \sum_{i=1}^{k}\sum_{j=1}^{m} P(x_{ij}) \log_k \frac{P(x_{ij})}{\pi_j P(x_i)}$$

$$= \sum_{i=1}^{k}\sum_{j=1}^{m} \pi_j P(x_i|S_j) \log_k \frac{\pi_j P(x_i|S_j)}{\pi_j P(x_i)} \quad \textbf{(11)}$$

where $P(x_i \mid S_j)$ is the conditional probability of finding a symbol $a_i$ given a sequence $S_j$. We expect high variance of $P(x_i \mid S_j)$ if the system we monitor is deviated. On the other hand, if the system we monitor is stable, we expect that the probabilities of each symbol $a_k$ in different SVs are very close, and therefore both the $I$ and $D_{GJS}$ are close to zero.

Also described in [17], the $D_{GJS}$ in Eq. 11 can be analytically approximated by using the Taylor expansion as

$$D_{GJS} \equiv \sum_{i=1}^{k}\sum_{j=1}^{m} \pi_j P(x_i|S_j) \log_k \frac{\pi_j P(x_i|S_j)}{\pi_j P(x_i)} \simeq \sum_{i=1}^{k}\sum_{j=1}^{m} \frac{(\pi_j P(x_i|S_j) - \pi_j P(x_i))^2}{\pi_j P(x_i)(2\ln k)} \quad \textbf{(12)}$$

Let's take a close look at Eq. 12 with Figure 4.The $m$ SVs with $k$ states/symbols ($k$ = 3 in Figure 4) can be considered an $k$ by $m$ contingency table if we multiply each SV by its weight and $N$ (the total number of symbols in $m$ sequences). Then, the Eq. 12 can be expressed by the Chi-square statistic $\chi^2$ [17], [37] as defined in Eq. 13:

$$\chi^2 \equiv N \sum_{i=1}^{k}\sum_{j=1}^{m} \frac{(\pi_j P(x_i|S_j) - \pi_j P(x_i))^2}{\pi_j P(x_i)} \simeq 2N(\ln k) D_{GJS} \quad \textbf{(13)}$$

We can then rewrite Eq. 13 to obtain the expected $D_{GJS}$ as shown in Eq. 14:

$$D_{GJS} \simeq \frac{\chi^2}{2N(\ln k)} \quad \textbf{(14)}$$

Therefore, given a certain significant level $\alpha$, the number of SVs $m$, and the number of states $k$, we can derive an asymptotical approximate threshold for the $D_{GJS}$, the $D_{GJS|k,m}$ , as:

$$P\big(D_{GJS} \le D_{GJS|k,m}\big) \simeq F\big(2N(\ln k) D_{GJS|k,m}, df\big) \Rightarrow D_{GJS|k,m} \simeq \frac{\chi^2_{df,1-\alpha}}{2N(\ln k)} \quad \textbf{(15)}$$

where $F$ is the Chi-square cumulative distribution function given the degree of freedom $df = (k - 1)(m - 1)$. $P(D_{GJS} \leq D_{GJS|k,m})$ denotes the probability of the $D_{GJS}$ less or equal to the threshold $D_{GJS|k,m}$. The $D_{GJS|k,m}$ in Eq. 15 is used as the criterion to determine whether the system deviation is significant. In Figure 7, we provide an example that shows how the $D_{GJS|k,m}$ works as the thresholds in our proposed monitoring approach, given that we have series of SVs shown in Figure 3.

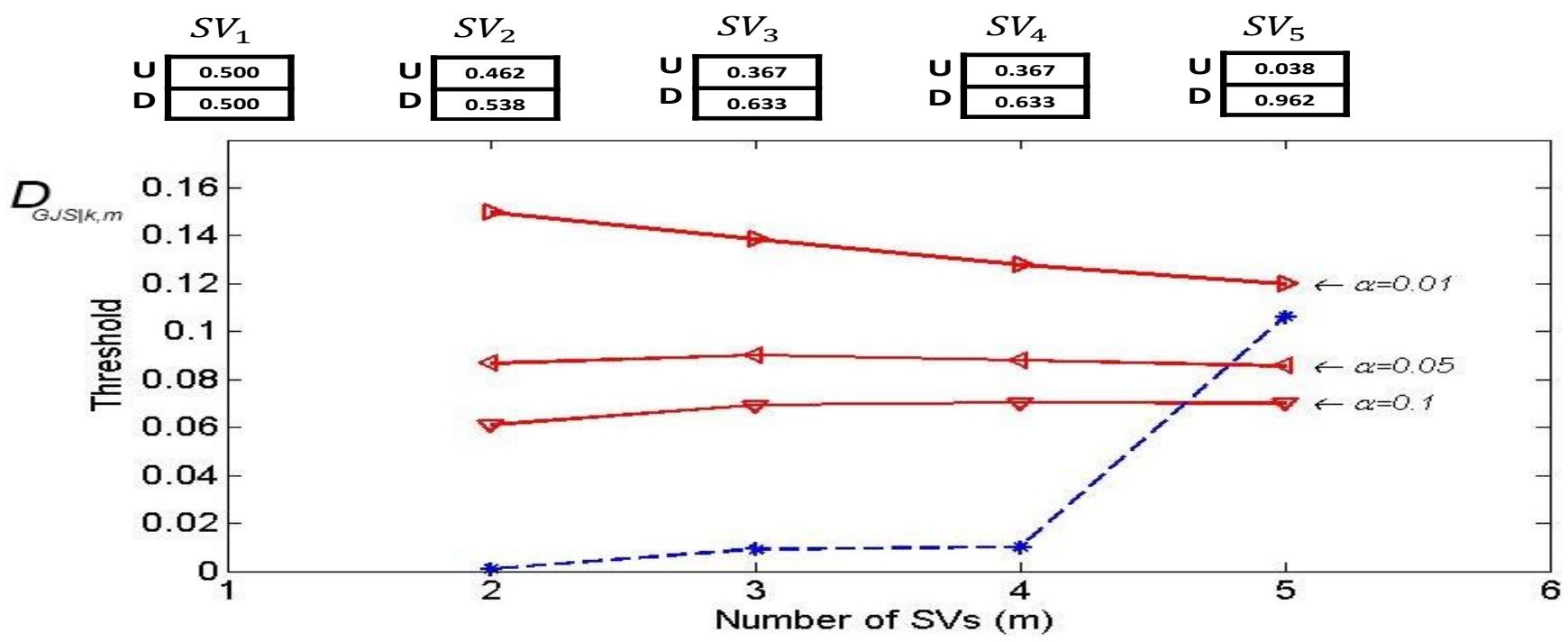

**Figure 7: A series of steady-state vectors with the significant thresholds**

Note that we consider all the SVs in Figure 7 are equally-weighted, which means, at the time when the system generates $m$ SVs, the $\boldsymbol{\pi}_1 = \boldsymbol{\pi}_2 = \ldots = \boldsymbol{\pi}_m = 1/m$. That is, for example, the weights for 3 SVs are all 1/3. Each SV is created from a sequence of 16 symbols as shown in Figure 3. The total length of all the sequences, $N$, increases as the system keeps converting the symbolic data stream into SVs. Note that $N$ must be sufficiently large to avoid obtaining the Chi-square statistic in Eq. 15 that may commit a Type II error. To calculate the $D_{GJS|k,m}$ when we have 2, 3, 4, and 5 SVs in Figure 7, for example, the $N$ is 2 * 16 = 32, 3 * 16 = 48, 4 * 16 = 64, 5 * 16 = 80, respectively. Thus the $D_{GJS|k,m}$ for 2, 3, 4, and 5 SVs in Figure 7 with $\alpha = 0.01$ are $\frac{\chi^2_{(2-1)(2-1),(1-0.01)}}{2*(2*16)*(\ln 2)} = 0.1496$, $\frac{\chi^2_{(2-1)(3-1),(1-0.01)}}{2*(3*16)*(\ln 2)} = 0.1384$, $\frac{\chi^2_{(2-1)(4-1),(1-0.01)}}{2*(4*16)*(\ln 2)} = 0.1279$, and $\frac{\chi^2_{(2-1)(5-1),(1-0.01)}}{2*(5*16)*(\ln 2)} = 0.1197$. In Figure 7 we provide 3 threshold lines for 3 different significant levels, $\alpha = 0.01$, 0.05, and 0.1. These 3 lines can also be interpreted as the probabilities of the $D_{GJS}$ higher than these lines, which are 0.01, 0.05, and 0.1. The lower the significant level $\alpha$, the higher the threshold $D_{GJS|k,m}$. Also in Figure 7, we can see the actual $D_{GJS}$ (dashed line), which compares $(SV_1, SV_2)$, $(SV_1, SV_2, SV_3)$, *and* $(SV_1, SV_2, SV_3, SV_4)$, are all lower than the threshold $D_{GJS|k,m}$. However, as expected, the $D_{GJS}$ at the time when we compare $(SV_1, \ldots, SV_5)$ is much higher than previous ones, as the $SV_5$ is significantly different from previous SVs. Given $\alpha = 0.05$ or $\alpha = 0.1$, the process monitoring system will give an alert to indicate that the system we monitor may deviate.

By modifying the aforementioned parameters of $D_{GJS}$, the proposed monitoring approach can not only be used in short-term change-point detections, but also gradual deviation monitoring. Figure 8 shows brief steps to use proposed process monitoring approach.

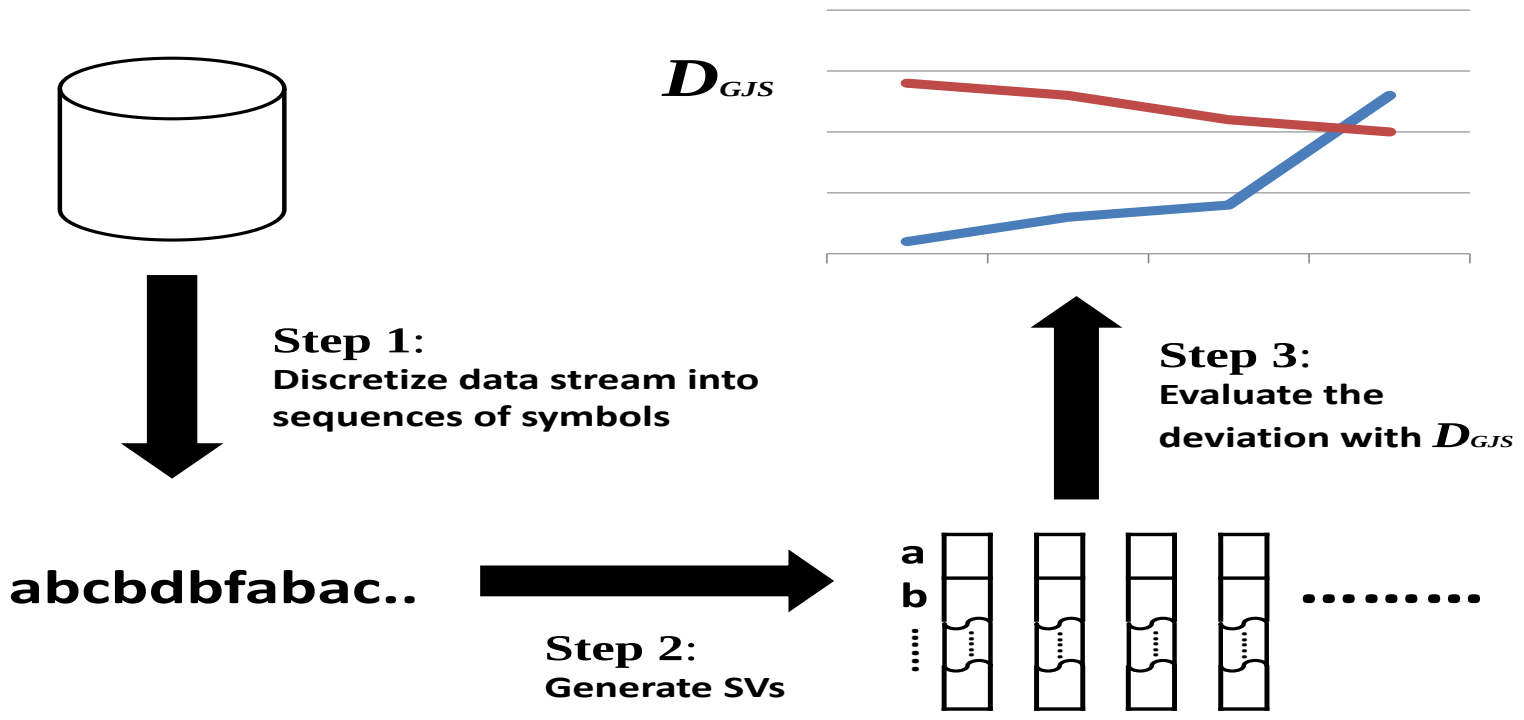

**Figure 8: Steps of sequence deviation assessment using $D_{GJS}$**

We first choose appropriate data discretization/reduction methods and the data granularity (the $k$ number of possible symbols that represent $k$ patterns of interest). The data stream is then symbolized into sequences. The next step is to calculate the stationary symbol probability distributions (SVs) in different times. We keep dividing all the symbols into $m$ equally-sized sequences of $n$ symbols. The length of each sequence depends on whether the sequence can well represent the wanted dynamics of the system we monitor in a period (e.g. a day/week/cycle). Then, $m$ by $k$ SVs are generated. In step 3, we compute the actual $D_{GJS}$ from SVs, and the significant thresholds of the $D_{GJS}$ given $k$ and $m$, with an appropriate significant level $\alpha$ (i.e. 0.1, 0.05, or 0.01). When the actual $D_{GJS}$ is higher than the threshold, the monitoring system gives an alert that indicates the system deviation is critically high and the deviation very unlikely occurs by chance. That is, the probability that the actual $D_{GJS}$ is higher than the threshold is the $\alpha$ we choose.

Again, note that the data discretization methods in Step 1 and the way to estimate the symbol probability distributions in Step 2 can be replaced by other approaches. In the beginning of Section 2, we briefly introduce the SAX, as we used it as an example in later experiments. However, many existing pattern classification methods and time-series representation techniques can be used in Step 1. We suggest using a data discretization method that can symbolize the data stream without losing too much information of interest. Also, the most intuitive way to estimate the symbol probability distribution is to count the numbers of occurrences of the symbols, the relative frequency vector (FV). However, we propose using SV instead here as it may capture unusual transitions of the system we monitor. In next section, we demonstrate the advantages of using proposed approach by comparing it to other measures.

## 4 Experiments

We investigate the applicability, limitation, and performance of the proposed approach by applying it to two different applications and comparing it to other existing sequence similarity/distance measures commonly used in the studies of sequence anomaly detections and the DNA/Protein sequence evolutions. Here, we first define two applications—the change-point and the long-term deviation detection for sequence data. The major differences between these two applications are the number of discrete sequences we compare and the types of changes we are interested in learning about. The change-point detection is about finding the significant high pairwise sequence distance, whereas the deviation detection is about evaluating the distance among multiple (more than two) sequences. We consider the change-point detection as the discovery of abrupt changes—difference of a system between two adjacent temporal sequences. On the other hand, the deviation detection is the notion of finding non-obvious evolutionary distance/relationships among multiple sequences—the gradual deviation of a system development. The experiments in this section include two aforementioned applications to both real-world and synthetic datasets. All experiments are implemented in R version 3.1.1 [38]. The datasets and source codes of distance/divergence measures used in the later experiments are available in [39] for readers to easily reproduce the experimental results.

### 4.1 Comparative Evaluation and Similarity Measures

To perform the comparative evaluation, we enumerate applicable distance measures from the literature about the sequence similarity analysis in different fields. Due to the fact that most of these measures are proposed to obtain the distance between two sequences, we consider the sum of the pairwise distance for each of them in comparison with $D_{GJS}$ that can calculate evolutionary distance among multiple sequences. The sum of pairwise distances is defined as:

$$D(S_1, S_2, \dots, S_m) = \sum_{i=1}^{m-1} D(S_i, S_{i+1}) \qquad \textbf{(16)}$$

where $m$ is the number of sequences we use to compute the distance. Also, a sliding window is used to keep generating the pairwise sequence distances, $D(S_x,S_y)$, of each paired sequences. As the actual changes/anomalies may occur anywhere in sequences, we use another fixed length monitoring window, denoted by $\Delta[i,j]$, to label whether anomalies occur within the period. Figure 9 shows the sliding window $W$ keeps shifting among sequences. The $D(S_x,S_y)$ denotes a calculated distance measure after we collect symbol data from sequences $S_x$ and $S_y$.

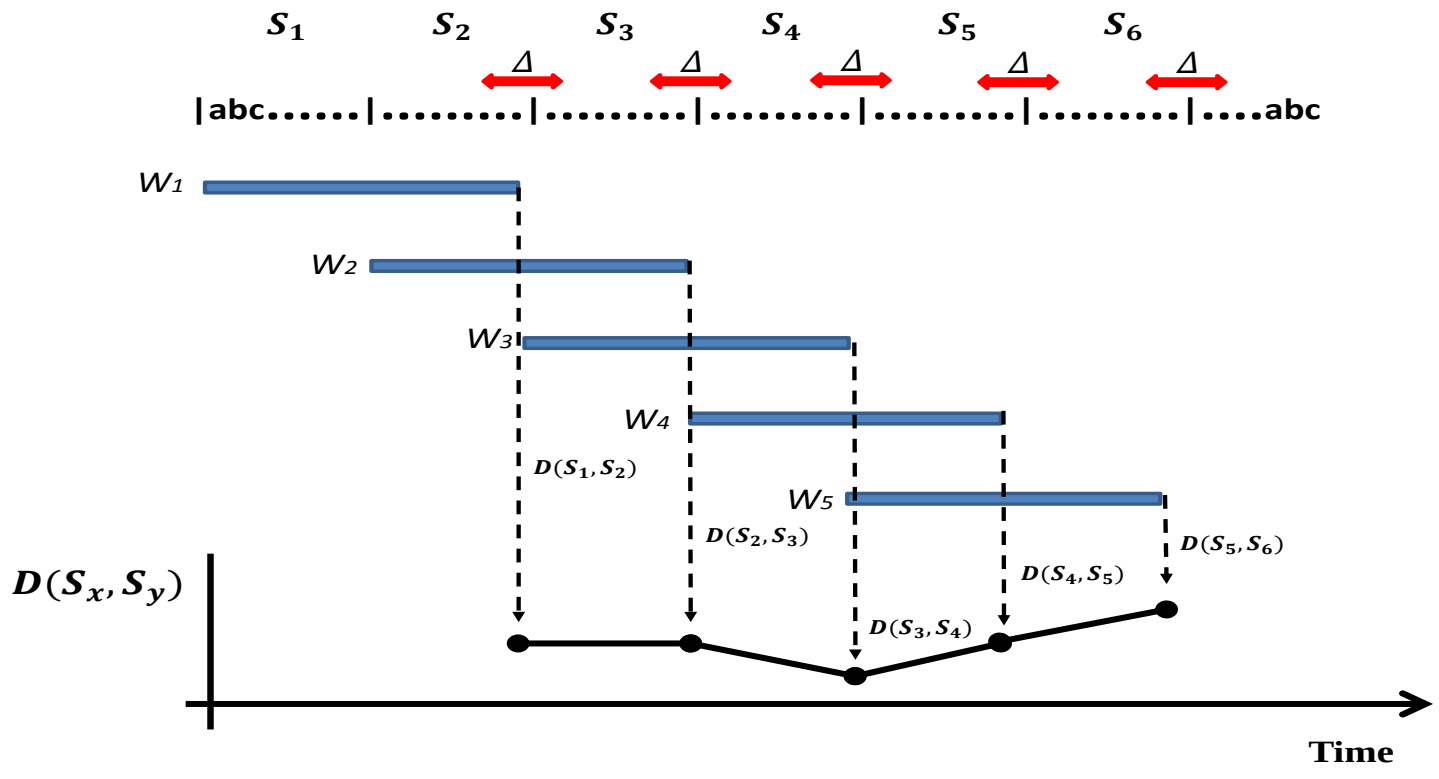

**Figure 9: Pairwise sequence distances**

Note that each Δ ranges beyond the boundaries of sequences. That is, we also consider a case of early warning that the anomalies may not only exist in the two sequences used to compute $D(S_x,S_y)$ but also may happen in the beginning of the next sequence. For example, $D(S_1,S_2)$ is computed after we have $S_1$ and $S_2$. However, the first Δ is across the boundaries between $S_2$ and $S_3$, as we believe that the anomalies could occur somewhere in the end of $S_2$. Besides, consider the application of deviation detection that takes more than two sequences into account to find the gradual changes, we use the sum of pairwise distances for each distance measure instead. Figure 10 shows an example that computes the gradual distance by comparing 3 sequences ($m$ = 3).

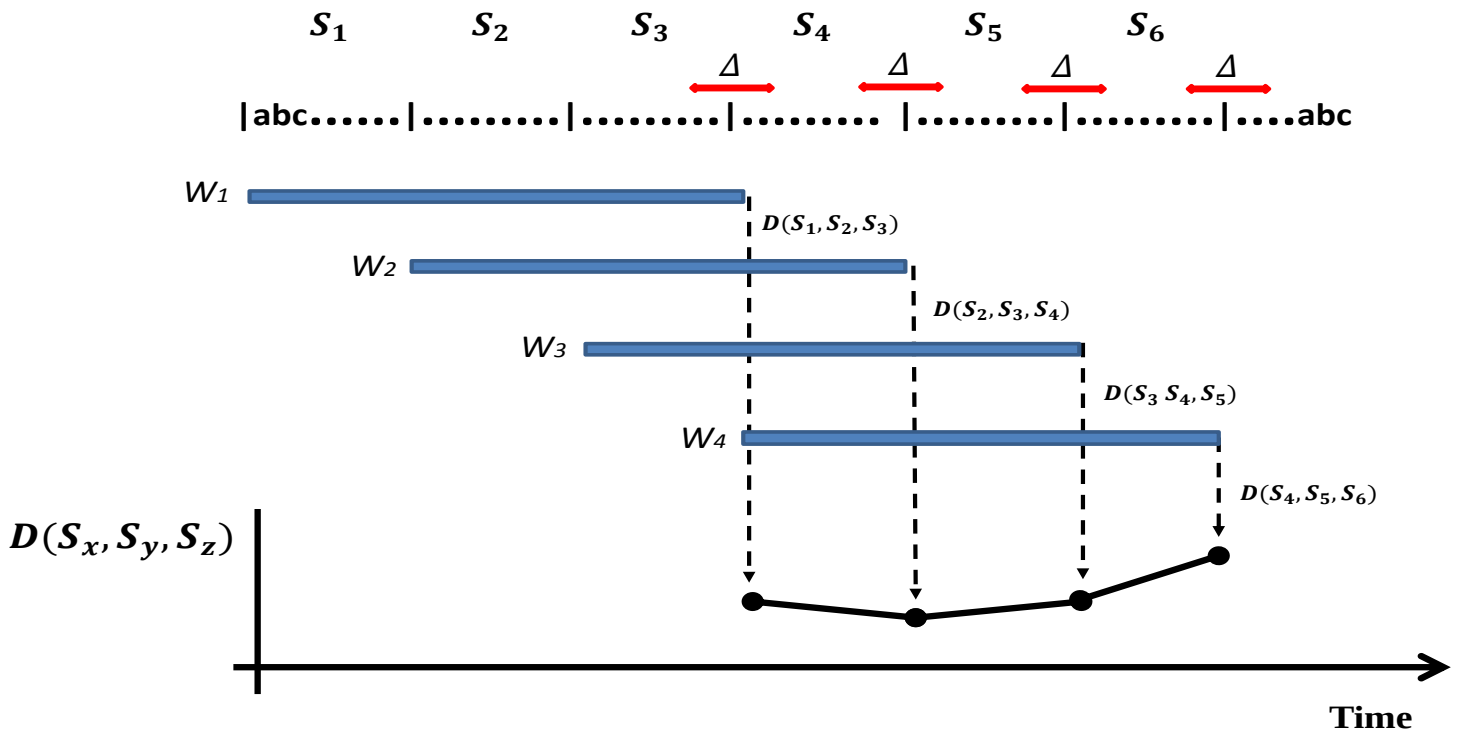

**Figure 10: Sum of pairwise sequence distances (given *m* = 3)**

Note that, for example, the first distance $D(S_1,S_2,S_3)$ is the sum of two pairwise distances, i.e. $D(S_1,S_2)$ + $D(S_2,S_3)$. Again, if there is any anomaly within a Δ, the target/outcome label of the corresponding distance measure will be positive (anomaly). These distances and labels are then used in the performance evaluation in the following subsections to create the ROC curves and compute the Area Under the ROC Curves (AUC).

As proposed approach is related to sequence similarity analysis found in various fields, in Table 1**,** we list the aforementioned divergences and five applicable distance measures with their notations used in later experiments. We choose these measures based on whether they can be applied to the comparison of the sequences with absent symbols. That is, those measures should allow for the comparison of two sequences in which a symbol that represents a system state may never occur. For example, consider two sequence $S_1$ (***a,b,c,a,b,c***) and $S_2$ (***a,b,a,b,a,b***). The symbol ***c*** is absent in $S_2$. The measure we use must be able to compute the distance/similarity that reflects the absentness of the symbol ***c***. Therefore, some of the measures, such as Paralinear distance [40] used in the calculation of distance of DNA/Protein sequences, are not applicable to our evaluation.

In Table 1, we define the *normalized length of Levenshtein distance* (*nLevD*) as Eq. 17, which is a measure that computes the ratio of edit distance (the number of insertions/deletions/substitutions operations needed to convert a sequence into another) between two sequences. $LevD(S_1,S_2)$ denotes the amount of edit distances between two sequences, whereas $|S_1|$ and $|S_2|$ are the length of sequences. We consider *nLevD* the degree of mismatch of two sequences and use it as a distance measure to detect the changes of a system. Similar to *nLevD*, *normalized length of the Longest Common Subsequence* (*nLCS*) [41] is a measure derived from the algorithm to find the *Longest Common Subsequence* (*LCS*). The *LCS* is a common but not necessarily consecutive subsequence among two or more sequences. It can be used to assess the similarity of sequences. In Eq. 18, $|LCS(S_1,S_2)|$ denotes the length of the longest common subsequence. The *nLCS* ranges from 0 to 1. The higher *nLCS* indicates higher similarity between sequences. In later experiments, we use $1 - nLCS$ as a distance measure and only consider the case of comparing two sequences, as finding the *LCS* for more than two sequences is an NP-hard problem [42] and thus impractical. Eq. 19 defines *Cosine Distance* as *1 - Cosine Similarity*. The Cosine Similarity is a common measure used to find the similarity between two documents in the field of text mining [43]. Instead, we use it to measure the similarity between two probability vectors, i.e. the discrete symbol probability distributions from the relative frequency vectors (FV) and the steady state vector (SV). In Eq. 19, $V(S)$ and $||V(S)||$ denote the symbol probability vector generated from a sequence and the norm of the vector respectively. As all the cells/components of the probability distributions (vectors) are greater or equal to 0, the Cosine Distance ranges between 0 and 1.

**Table 1**: Sequence similarity measures

| Measure | Notation | Equation |
|---|---|---|
| *Generalized Jensen-Shannon Divergence on Steady-State Vectors* | $D_{GJS}+SV$ | (Eq. 9) |
| *Generalized Jensen-Shannon Divergence on Relative Frequency Vectors* | $D_{GJS}+FV$ | (Eq. 9) |
| *Kullback-Leibler Divergence on Steady-State Vectors* | $D_{KL}+SV$ | (Eq. 4) |
| *Kullback-Leibler Divergence on Relative Steady-State Vectors* | $D_{KL}+FV$ | (Eq. 4) |
| *Normalized length of Levenshtein distance* | ***nLevD*** | $nLevD(S_1,S_2)=\frac{LevD(S_1,S_2)}{\sqrt{\lvert S_1\rvert * \lvert S_2\rvert}}$ **(17)** |
| *One minus Normalized length of the Longest Common Subsequence* | ***1-nLCS*** | $1-nLCS(S_1,S_2)=1-\frac{LCS(S_1,S_2)}{\sqrt{\lvert S_1\rvert * \lvert S_2\rvert}}$ **(18)** |
| *Cosine Distance on Steady-State Vectors* | ***CosDist + SV*** | $CosDist(S_1,S_2)=1-\frac{V(S_1)\cdot V(S_2)}{\lVert V(S_1)\rVert * \lVert V(S_2)\rVert}$ **(19)** |
| *Cosine Distance on Relative Frequency Vectors* | ***CosDist +FV*** | (Eq. 19) |
| *p-Distance* | $D_p$ | $D_p(S_1,S_2)=\frac{d}{n}$ **(20)** |
| *Jukes-Cantor distance* | $D_{JC}$ | $D_{JC}(S_1,S_2)=\begin{cases}-\frac{k-1}{k}\ln\left(1-\frac{k}{k-1}D_p\right), \text{if } D_p\le\frac{k-1}{k}\\ +\infty, \text{if } D_p>\frac{k-1}{k}\end{cases}$ **(21)** |

As discussed in previous sections, many similarity/distance measures are proposed to help find the evolutionary distances of DNA/Protein sequences. In later experiments, we consider two applicable measures—the *p-distance* ($D_p$) and *Jukes-Cantor distance* ($D_{JC}$) [44] as defined in Eq. 20 and Eq. 21. The $D_p$ is the proportion of locations that differ between two sequences. In Eq. 20, $d$ is the number of one-to-one mismatched symbols and $n$ is the length of a sequence. Note that the lengths of two sequences must be the same. The $D_p$ is simple and easy to compute, but it underestimates the possible substitution of each symbol at each location. Consider three possible symbols (***a***, ***b***, ***c***) in a sequence as an example. Each symbol in the sequence can be replaced by two other symbols. That is, for $k$ possible symbols in a sequence, each symbol in a sequence can be replaced by $k$ *-1*. The p-distance does not reflect the granularity of the sequence data that contributes to the distance of two sequences. The $D_{JC}$ is proposed to correct the problem. It is originally assumed that the nucleotide symbol substitution rate (replacement rate) and symbol frequency are all equal. It is also applicable in our experiments, as all symbols are assumed equal-probable before the discretization. In Eq. 21, $k$ is the number of possible symbols (states). Note that, by the original definition, the $D_p$ in Eq. 21 is expected to be smaller than ($k$-1)/$k$. Instead, we use +∞ when the $D_p$ is higher than ($k$-1)/$k$.

### 4.2 Detection of Sequence Change Points

As the accuracy of anomaly labels are difficult to verified and obtained [1], for illustration purpose, we first created two synthetic datasets called DC (which denotes “Distribution Change”) and JM (which denotes “Jumping Mean”), and discretized them into temporal sequences. The DC dataset is created as follows. We randomly generate 900 uniformly-distributed data points between -3 and +3, followed by another 300 normally-distributed random data points with the parameters ($\mu$ = 0, $\sigma$ = 1). This process is repeated 25 times and all generated data are then concatenated to form the DC dataset with 30,000 data points. Figure 11 shows the first 3,600 data points of the DC dataset.

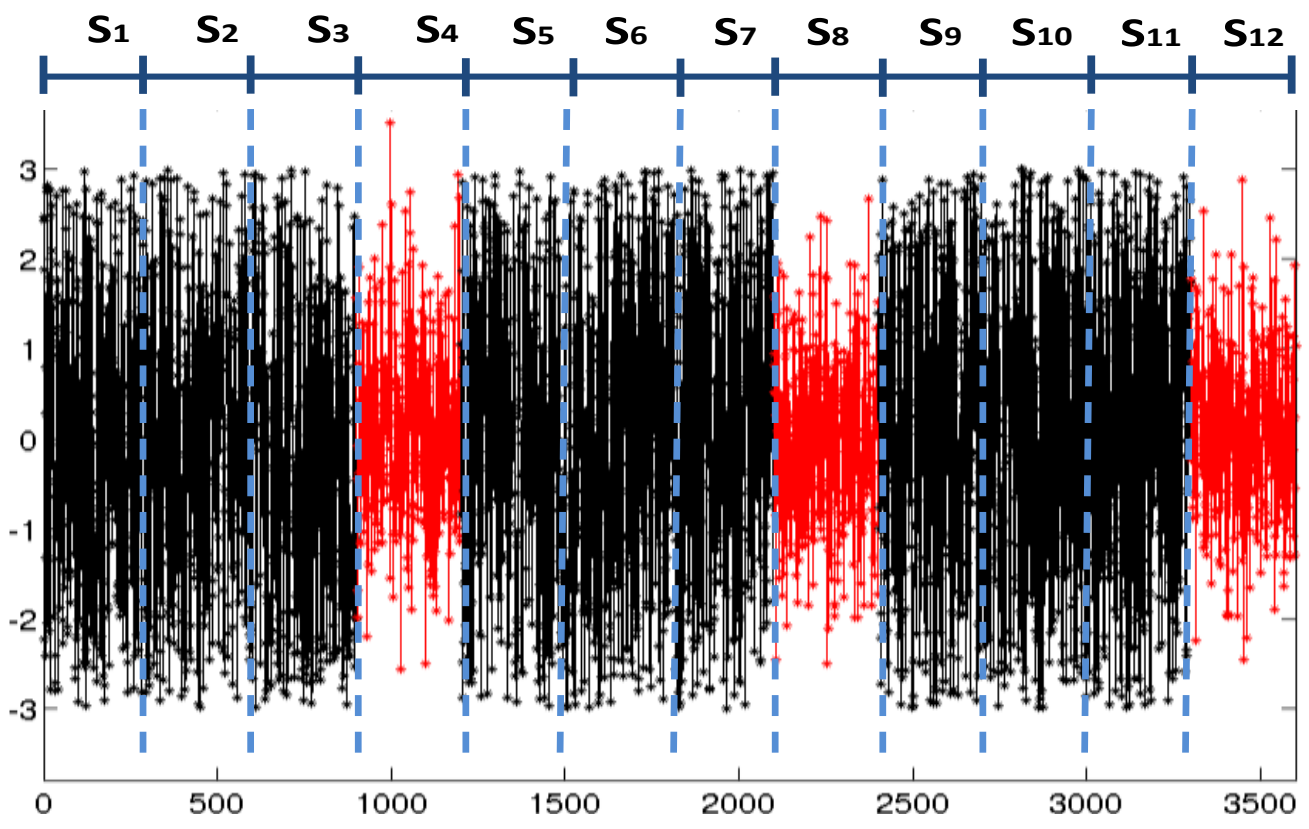

**Figure 11: First 3,600 data points from DC dataset**

The changes of data distributions (anomalies) in the raw series data are obvious. They can be easily detected by various detection techniques, such as likelihood ratio test. However, the goal here is to reduce/compress the raw data by discretization methods and see whether the methods based on aforementioned sequence similarity measures can also detect the differences among discrete symbol sequences. We symbolize DC dataset using the SAX with different numbers of possible symbols ($k$). The segment size to create a symbol in the SAX is 3 data points, i.e. there are a total of 10,000 symbols generated from the DC dataset. The length of each sequence, $n$, is set to 100 symbols. Therefore, for example, there are 12 sequences (1,200 symbols) from the data points in Figure 11. Apparently, the change points are at 901, 1201, 2101, 2401, and 3301 (i.e. at the beginning of the sequence $S_4$, $S_5$, $S_8$, $S_9$, and $S_{12}$). The pairwise $D_{GJS}$ and other distance measures $D(S_x,S_y)$ are then computed. As we calculate the distances after receiving pairs of sequences, for total $m$ sequences we can obtain ($m$ – 1) distances. Figure 12 shows the actual pairwise $D_{GJS}$ in Figure 11.

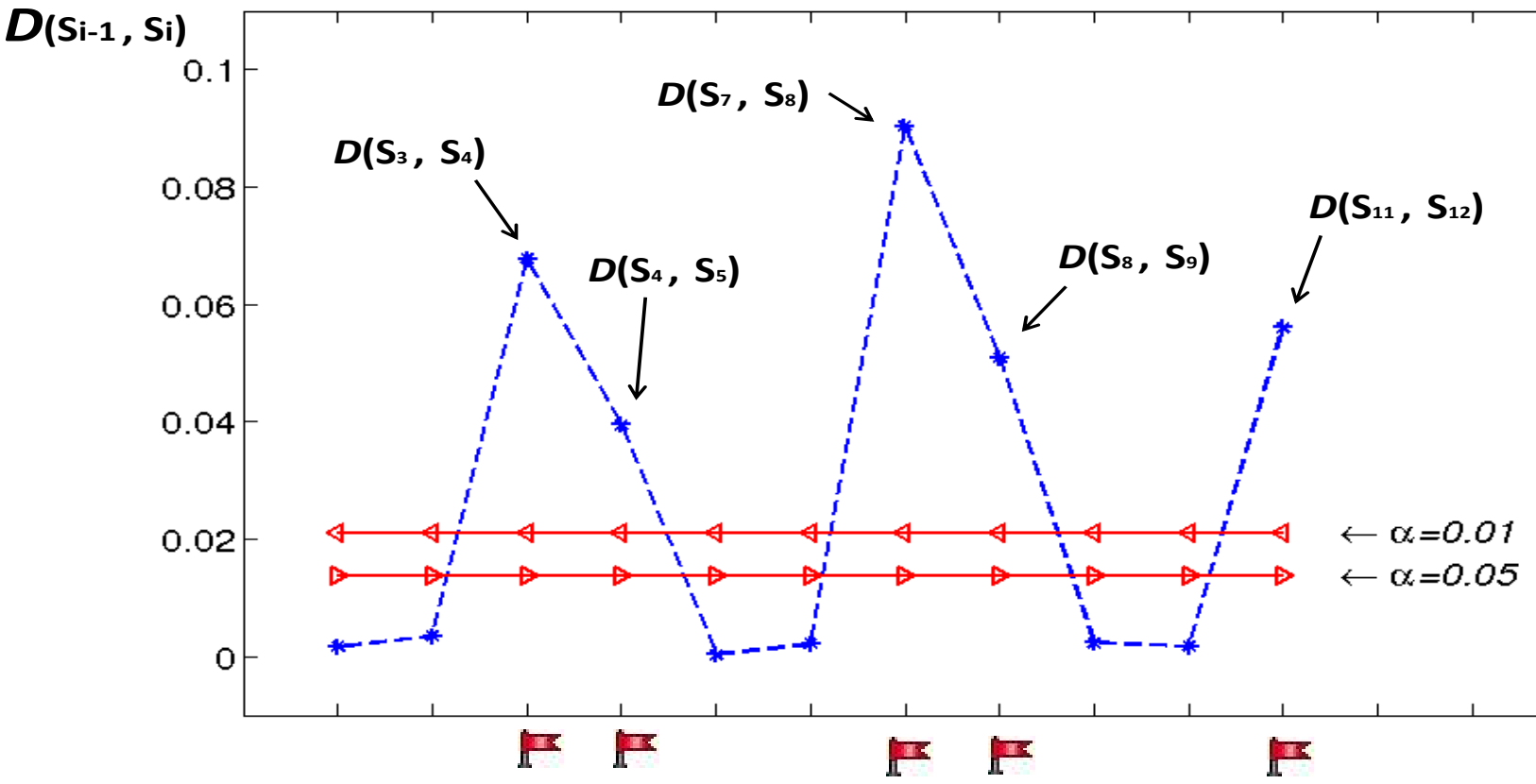

**Figure 12: $D_{GJS}$ and the thresholds with different $\alpha$**

Note that the number of possible symbols ($k$) in Figure 12 is 3, and 11 pairwise $D_{GJS}$ are computed. In our approach, the significant threshold $D_{GJS|k,m}$ is used to help identify the changes. Figure 12 also shows the thresholds with different $\alpha$ (0.05 and 0.01). The thresholds in different times are all the same (which are $\frac{\chi^2_{(3-1)(2-1),(1-0.05)}}{2*(2*100)*(\ln 3)} = 0.0136$ and $\frac{\chi^2_{(3-1)(2-1),(1-0.01)}}{2*(2*100)*(\ln 3)} = 0.0210$ for $\alpha = 0.05$ and $\alpha = 0.01$ respectively), because we continuously compare two sequences. The number of sequences we compare ($m$ in Eq. 9) and the total number of symbols in two sequences ($N$ in Eq. 9) are thus always 2 and 200. Provided that the significant level $\alpha = 0.05$, we can see the $D_{GJS}$ are higher than the thresholds at the times when two sequences from different distributions are compared. The monitoring system based on our approach can thus raise the red flags and alert us for the changes.

As all the scales of the aforementioned measures are not equal, we consider using the AUC to compare our approach to other distance measures. For DC dataset, the $\Delta$ is set to [-5, +5] at the end of each sequence, which certainly can capture the change points and generate the positive anomaly labels used in plotting ROC curves. Figure 13 shows the AUCs of the measures with different number of possible symbols ($k$). It suggests that those measures based on finding the (mis)matched symbols perform poorly compared to those based on computing the distances of two symbol probability distributions. One major reason is that the DC dataset is randomly generated. In this case, the proportion of matches between two sequences is usually lower. Besides, we can see that the AUCs of most of the measures (except $D_{GJS}$) decreases as the number of possible symbols ($k$) increases, which also indicates that the performance of these measures declines when they are applied to high-granularity temporal sequence data. Apparently, they should not be used as the measure in the process monitoring. On the other hand, the advantages of using $D_{GJS}$ and SV are clear. Even with higher $k$, the AUCs of $D_{GJS}$ are nearly constant when $D_{GJS}$ is used in the comparison of two ($m$ = 2) sequences. Also, the $D_{GJS} + FV$ performs slightly better than $D_{GJS} + SV$, as FV is a better estimate than SV when they are both applied to measuring the symbol probabilities from data points randomly generated from a given distribution. Another interesting result is that the AUCs of $D_{KL} + SV$ is higher than the AUCs of $D_{KL} + FV$, which suggest that using SV can improve the performance of $D_{KL}$ when $D_{KL}$ is used in measuring the differences of higher-granularity sequence data.

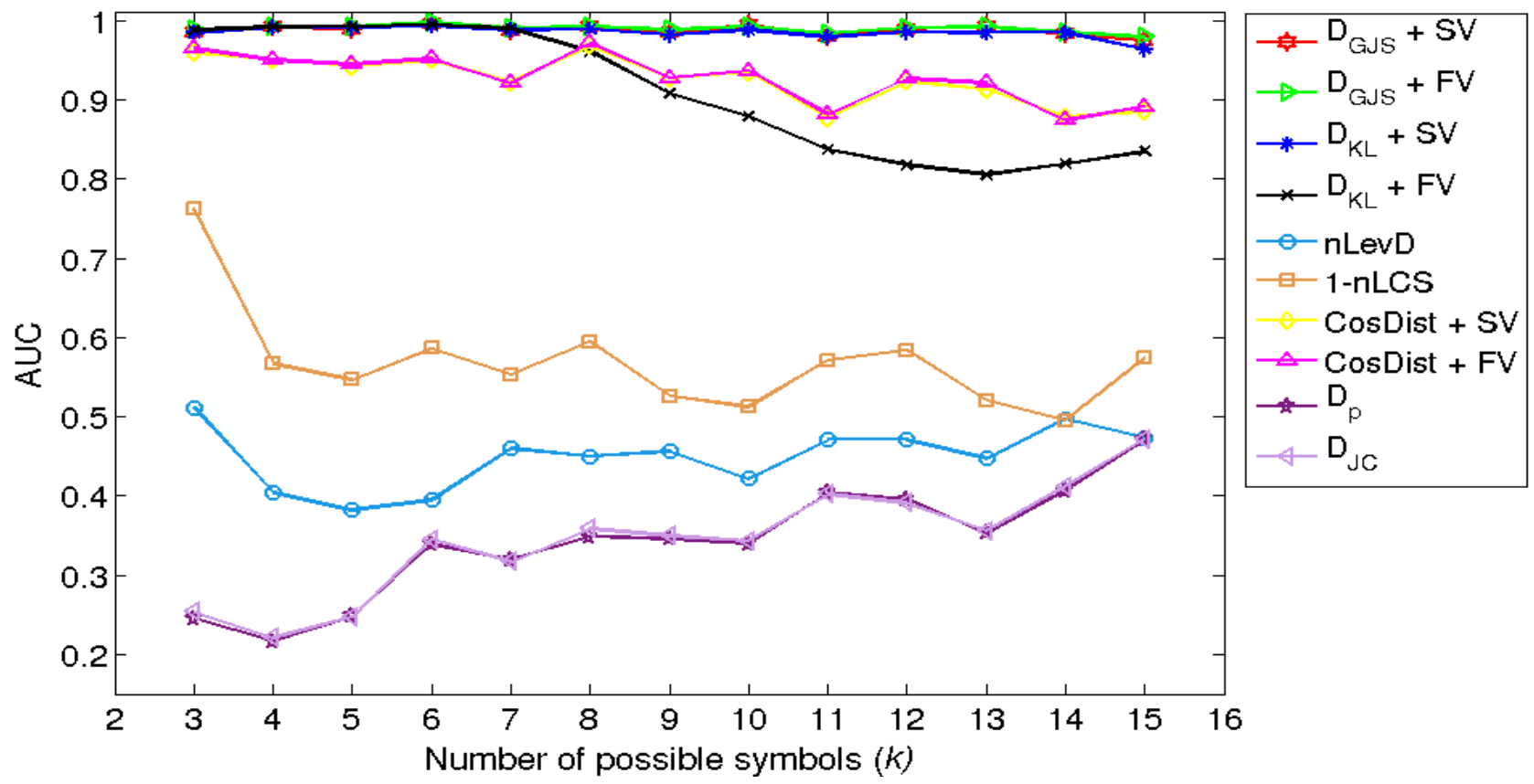

**Figure 13: The AUCs of distance measures for DC dataset**

Consider a different application to detect jumping means. The JM dataset shown in Figure 14 consists of 30,000 data points generated by the following auto-regressive model borrowed from [5].

$$X_t = 0.6\,X_{t-1} - 0.5\,X_{t-2} + \varepsilon_t$$

where $\varepsilon_t$ is the Gaussian random variable with mean $\mu$ and standard deviation $\sigma = 1$. The change points are inserted at time 1,000$x$ ($x = 1, 2, \ldots, 29$). The mean $\mu$ at time $t$ is defined as:

$$\mu_t = 3\left\lfloor\frac{t}{1000}\right\rfloor$$

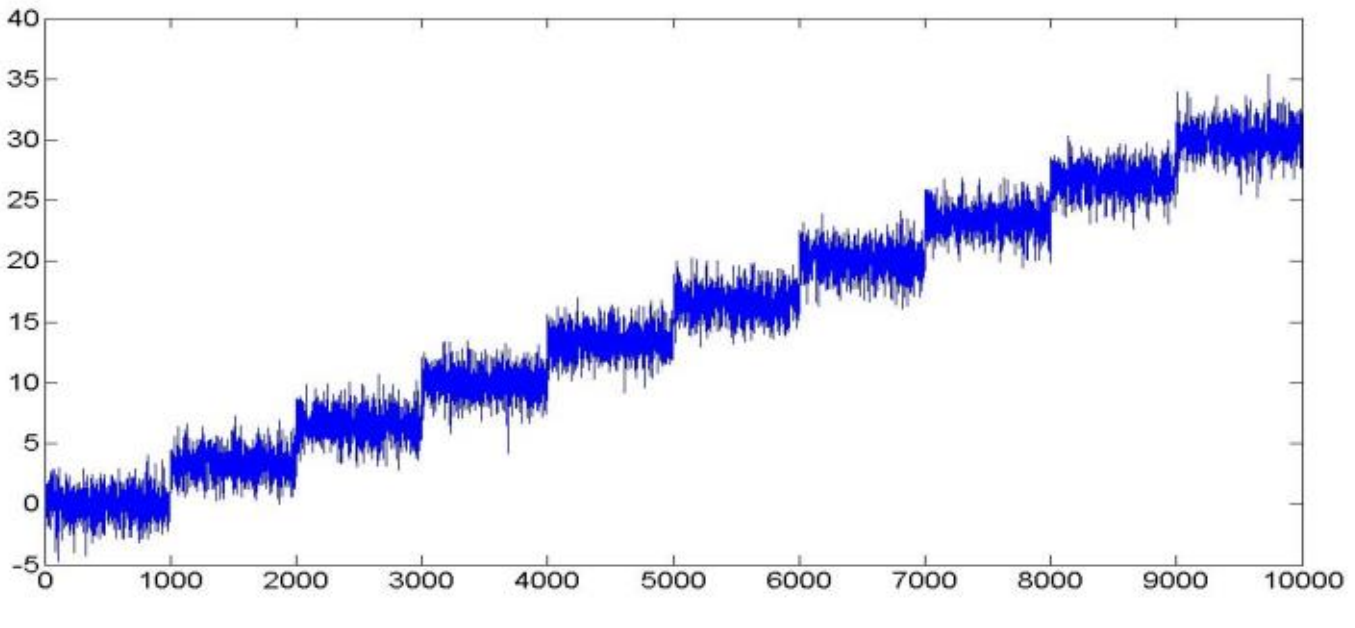

**Figure 14: First 10,000 data points from JM dataset**

Obviously, the change points in the JM dataset without any data discretization can be easily detected by those techniques based on monitoring the shifting means (e.g. CUSUM). However, we would like to know whether a monitoring system that uses measures in Table 1 can also detect this type of changes in discrete sequences generated from a dataset like JM dataset after performing data discretizations. Again, we discretize raw data by using SAX given different numbers of possible symbols ($k$). The segment size to create a symbol is set to 2 data points, and the length of each sequence, $n$, is set to 50 symbols. Figure 15 shows the AUCs of the measures with different numbers of possible symbols ($k$) for JM dataset. Due to the nature of SAX, we expect that higher $k$ (higher granularity) for SAX will lead to higher AUCs. Figure 15 also shows that the $D_{GJS}$ is a better measure. Also, with higher granularity sequence (higher $k$), using $D_{KL}$ on SV can improve the performance of sequence anomaly detection in terms of AUC.

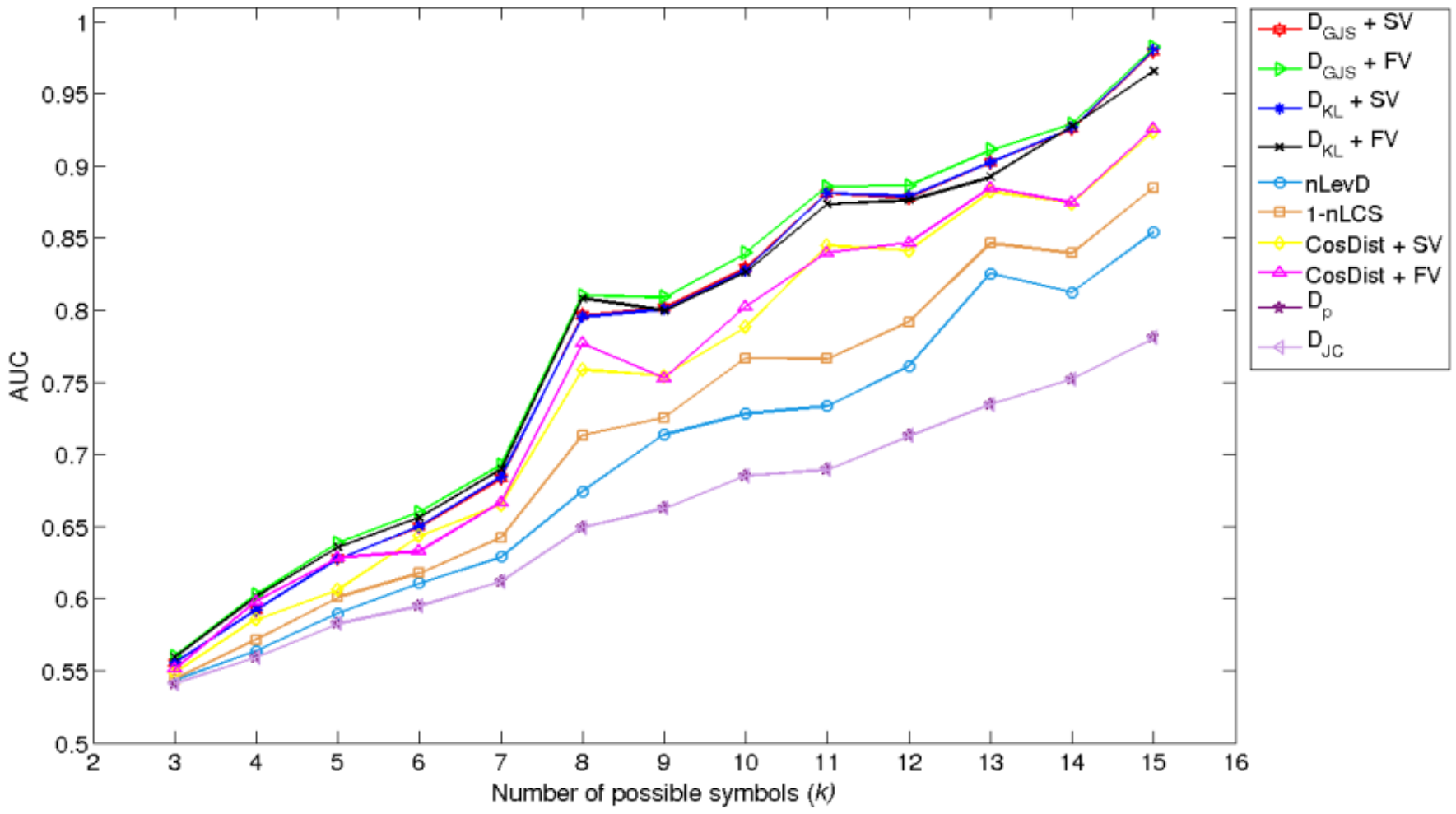

**Figure 15: The AUCs of distance measures for JM dataset**

The third dataset we used is a real-world dataset about the detection of ozone level and is publicly available at UCI Machine Learning Repository website [45]. As discussed in [46], high ground level ozone is potentially harmful to human health. The dataset, "eight hour peak set" (eighthr.data) from 1998 to 2004, contains some features that might be useful to the identification of the ozone/normal days. As most of these features may be irrelevant [46], we use the "ozone days" class labels in the dataset and only select the wind speed variables in different times/hours (WSR0 to WSR23) for simplicity. There are some missing values in the raw dataset, especially in the second half of the year 2002. So, we only use the data before the second half of 2002 in the later experiment. Then, we perform principal component analysis and consider the first principal component score for the experiment, as it accounts for most of the variability of the wind speed dynamics. Figure 16 shows the scores with the ozone day labels denoted by red dots below.

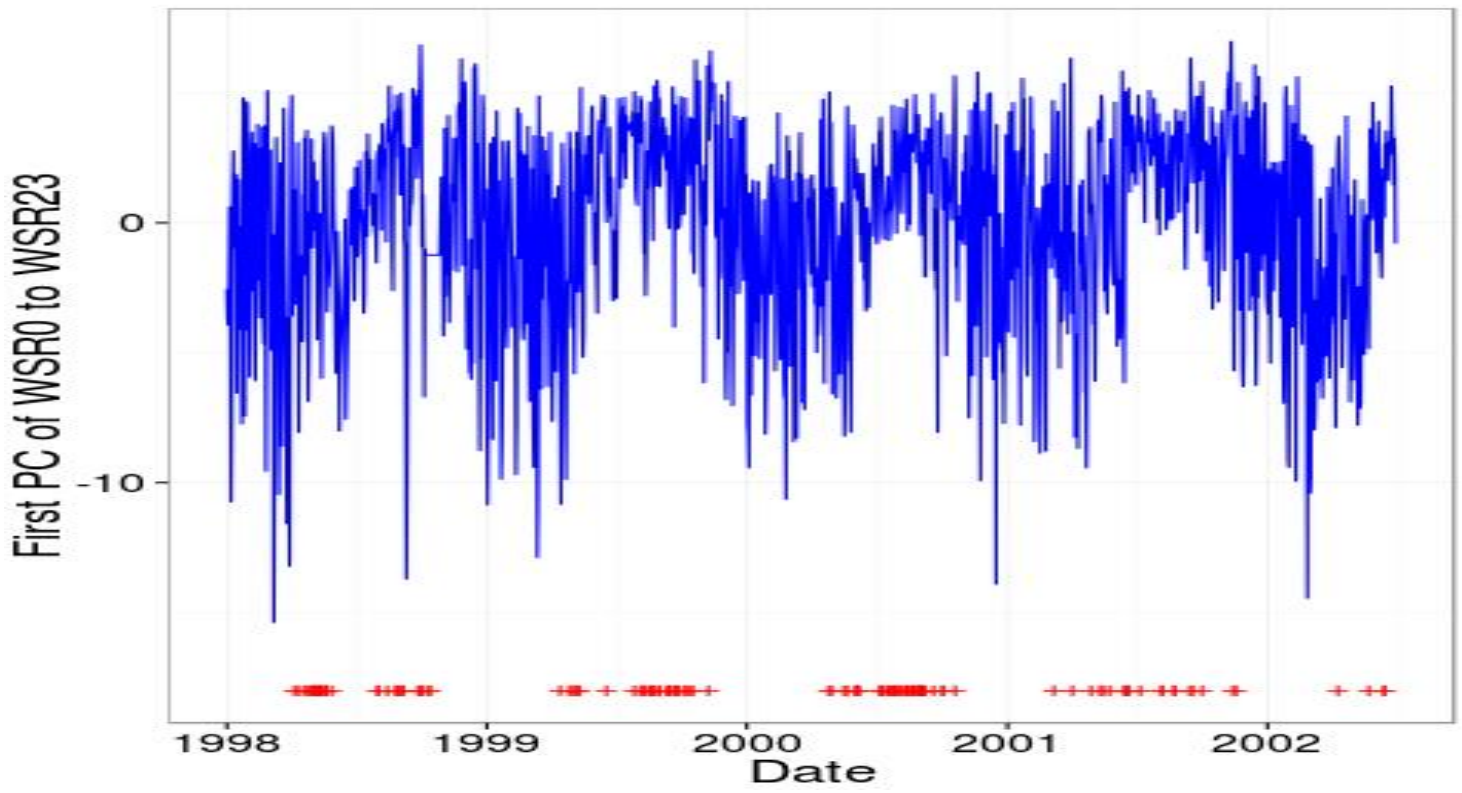

**Figure 16: Wind speed 1st PC score and ozone day label**

The goal of using the sequences generated from the series data shown in Figure 16 is to find out whether the proposed approach applying to the wind speed pattern dynamics can help detect the change points between ozone and normal days. Again, we discretize the data using SAX with different numbers of possible symbols *k*. The segment size to create a symbol is set to 1 data point per day (i.e. no numerosity reduction), and the length of each sequence, *n*, is set to 14 symbols/days. The $\Delta$ is set to [-7, +1] at the end of each sequence, which means we experimentally consider that the change points may occur anytime in the last 7 days/symbols of two sequences we compare, or in 1 day before it actually happen as an early warning. We present the AUCs of the aforementioned measures with different *k* in Figure 17.

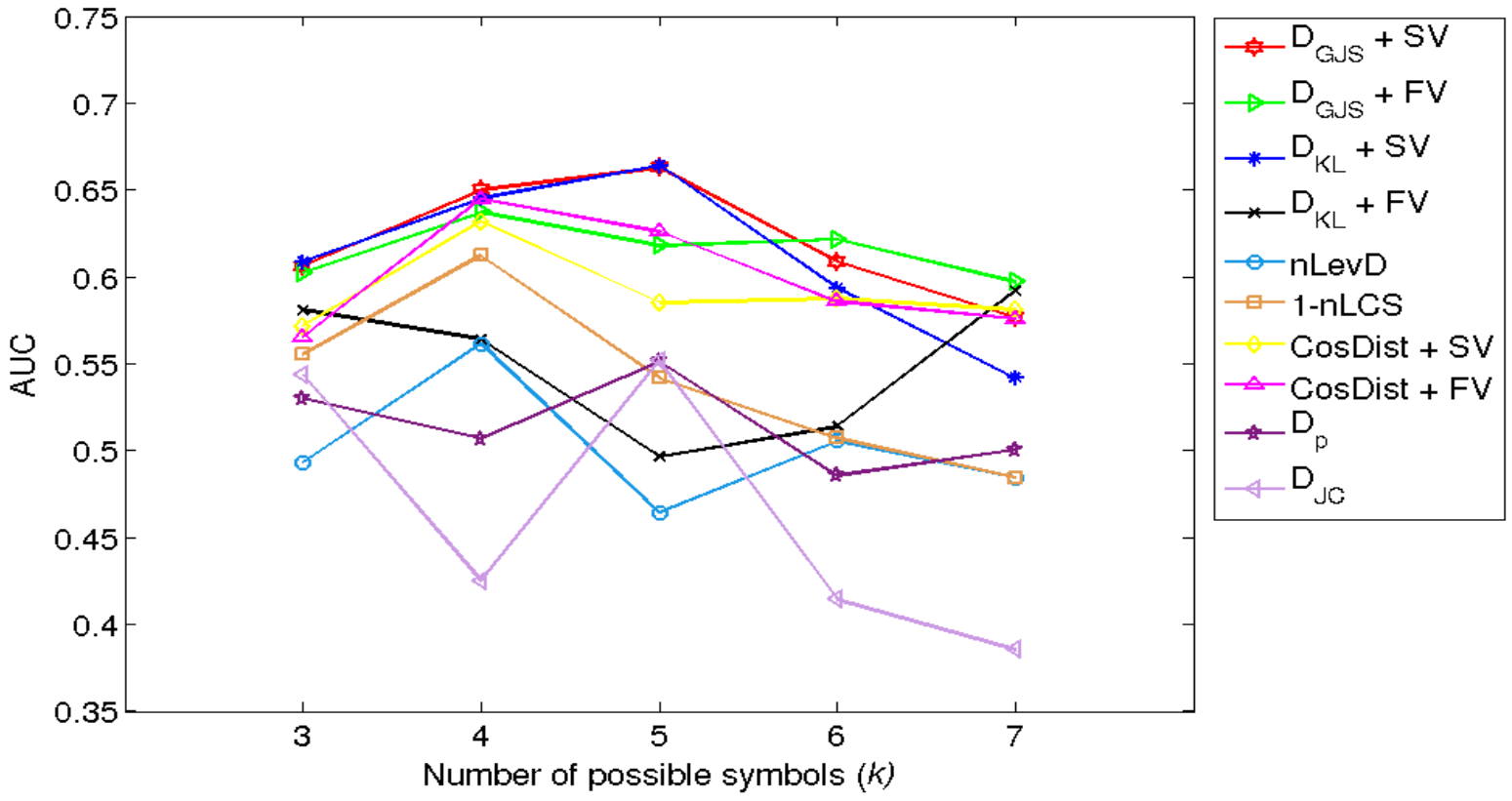

**Figure 17: The AUCs of distance measures for wind speed 1st PC score**

Note that we consider *k* between 3 and 7 to avoid resulting in too many absent symbols, as the size of each sequence, *n*, is 14 symbols/days. Figure 17 shows that $D_{GJS}$ perform relatively well, and using SV instead of FV may improve the performance of $D_{KL}$ in terms of AUC. We can also see that the performance of these measures decrease (and become useless in terms of AUC) as the $k$ increase, partly because $n$ we chose is too small to characterize the symbol transitions. It is always a difficult problem to choose an appropriate size of a sequence to accurately detect the anomalies. Longer sequence may result in the delay of the alarm, whereas a shorter sequence may not be able to discover the symbol transition patterns. Besides, we tried to identify change points between ozone and normal days just by monitoring the sequences of wind speed pattern dynamics without the ozone day class label used in learning the distances/similarities among sequences, which is different from the applications using supervised learning algorithms originally presented in [46].

### 4.3 Detection of Sequence Deviation

We here consider a different application to monitor long-term deviations. The data we used is a real-world dataset collected by power stations on the border between Croatia and Bosnia. Those stations in different locations recorded the measurement (Megawatt Hour, MWh) of the power transmission/consumption every 15 minutes from 2005 to 2008. We select one dataset from an active station named CAF_BIH [39], which consists of 137,568 data points. The goal is to see how aforementioned measures can identify the deviation of the power usage development, the sequential pattern changes (symbol transitions), to help detect the power surges/spikes (the anomalies). As the power surges are expected to be rare, we use cut-points to discretize the CAF_BIH data instead of using SAX. We first consider any data points greater than 20 MWh as the power surges as shown in Figure 18. Then, the cut-points are used to discretize the data and determine the number of possible symbols ($k$). In Figure 18, we provide two cut-points (the dashed lines at 10 and 20) as an example that discretizes the CAF_BIH data into a symbolic data stream with $k = 3$. That is, 137,568 data points become a long discrete sequence consisting of 3 possible symbols (***a***, ***b***, ***c***), based on which area a data point is in.

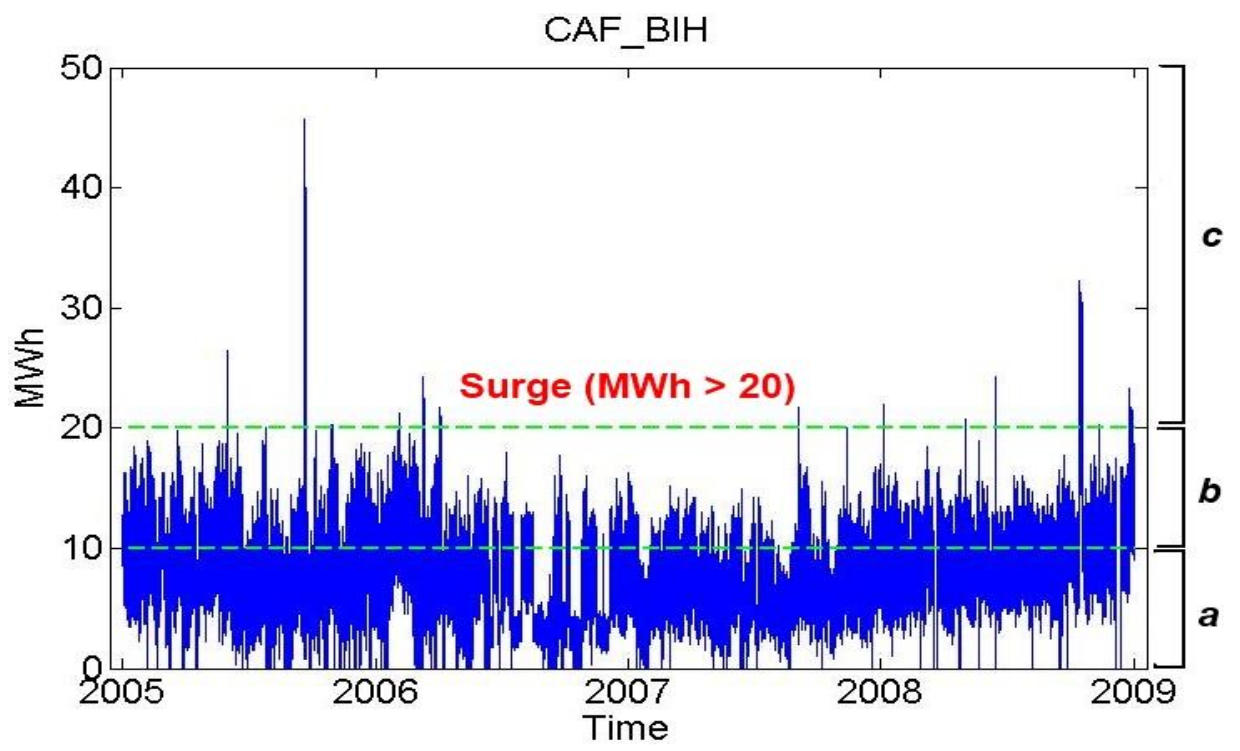

**Figure 18: CAF_BIH power consumption data**

The next step is to determine the size (length) of each sequence ($n$) to compare. Again, a long sequence may result in the delay of the alarm, whereas a short sequence may not be able to discover the symbol transition patterns (e.g. the cycle of the symbol transitions). We empirically choose 48 symbols ($n$ = 48), which is equal to 12 hours, as the length of for each sequence. Figure 19 shows a sample of the settings used in our experiments on the CAF_BIH data. Note that the Δ is set to [-48, +4] (52 symbols), which is minus 12 hours/plus 1 hour at the point when the distance measures are computed. In this case, the surges occurring within this range will be captured and used as the positive anomaly labels in the later ROC analysis to calculate the AUCs. Again, the length of the Δ could vary among different applications. In Figure 19 and all of the experiments for the CAF_BIH data, we consider the case that we have been monitoring the power usage development for a certain period (e.g. $m$ = 3, sequences = 36 hours in Figure 19). We expect that the surges may occur in the last sequence (last 12 hours) or in the near future (after 1 hour) as the early warnings.

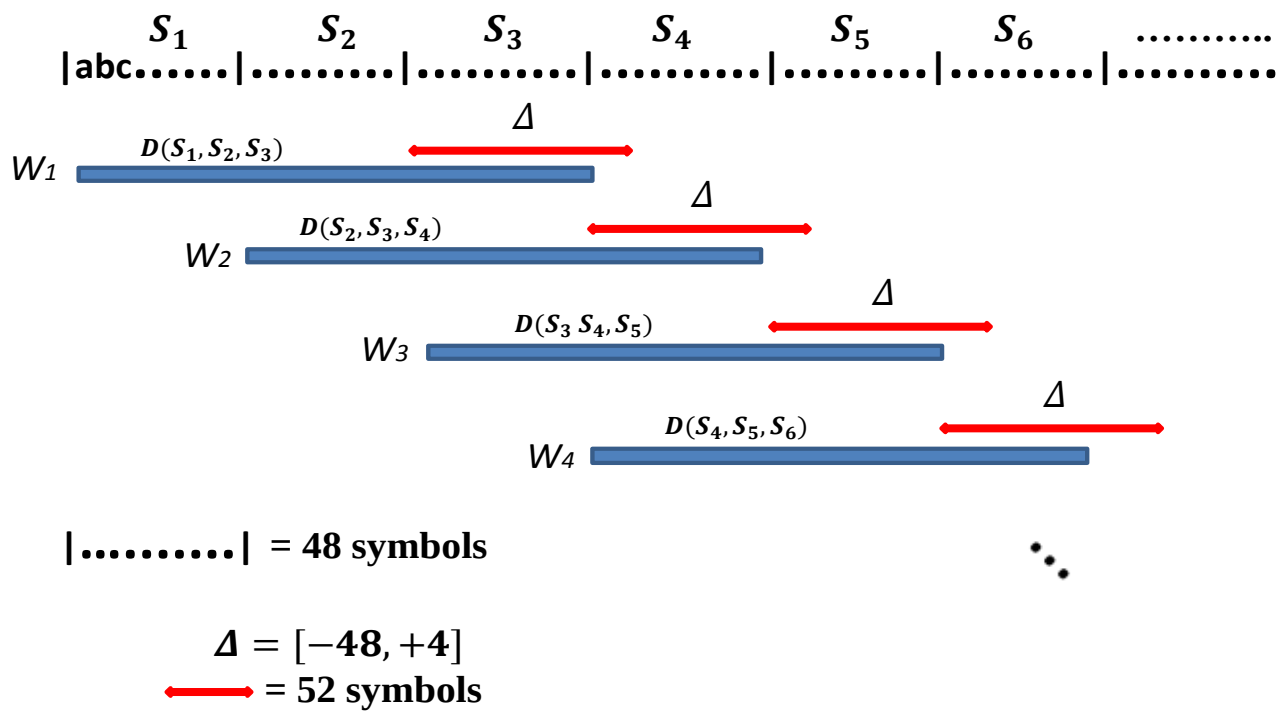

**Figure 19: Sequences with sliding windows ($m$ = 3) for CAF_BIH**

To discretize CAF_BIH data with different $k$, we use different numbers of cut-points to divide the data below the surge line (MWh <= 20) into equally-sized areas marked as symbols. Those data points above the surge line are also labeled as surges/anomalies. Figure 20 shows that the measures based on comparing symbols probability distributions from Steady-State Vectors (SV) perform better than those based on counting the ratio of (mis)matches of symbols do, especially when all measures are applied to the comparisons of multiple sequences to obtain the evolutionary distances.

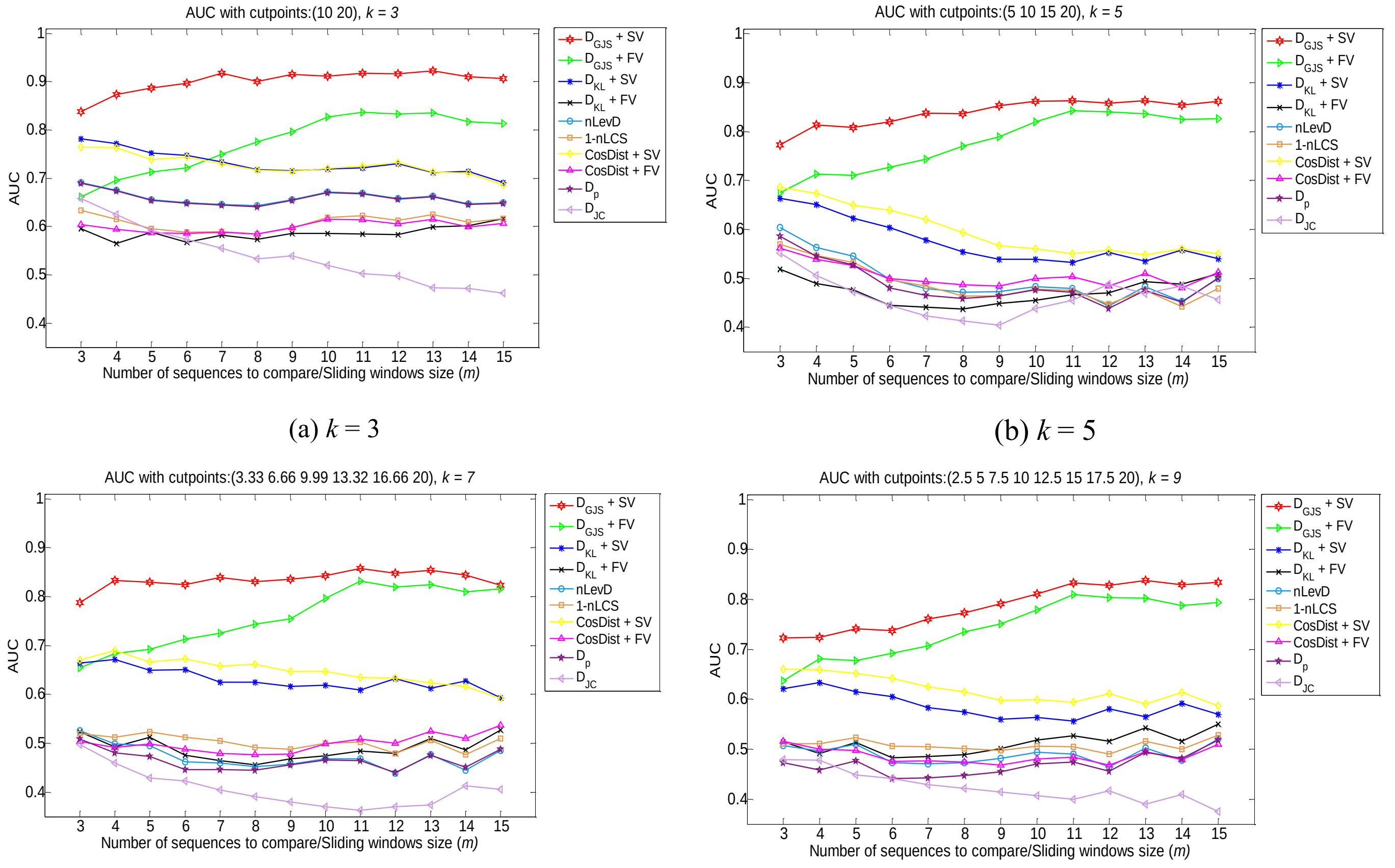

(a) $k = 3$ (b) $k = 5$

(c) $k = 7$ (d) $k = 9$

**Figure 20: AUCs with different parameters for CAF_BIH data**

In addition, we suggest using $D_{GJS}$ instead of $D_{KL}$, as we observe that the generalizable and symmetric divergences are better metrics to assess the deviation of a system development. Figure 20 shows most of AUCs for $D_{GJS}$ are higher than those for $D_{KL}$. Also, we can see that using SV results in higher AUCs for $D_{GJS}$, $D_{KL}$, and *CosDist* in most cases, as SV can maximize the differences among the symbol probability distributions we compare. On the other hand, using FV performs comparably well, especially when we compare more sequences. It can be explained by the nature of SV and FV. As discussed, the SV can be considered a snapshot of a system development. The comparison of fewer sequences can maximize the system deviation within a smaller time frame (the length of the all sequences we compare). Nevertheless, when more sequences are used in the comparison (higher *m*), the measures that use FVs might collect enough information about symbol probabilities close to the real symbol probability distribution. Also, we assume that using SV to estimate symbol probabilities is better than using FV in most cases, because SV can also capture the unusual symbols transitions. However, in a special case that there are only a few transitions in a long sequence, FV might be a better estimator, since FV represents the actual symbol probabilities but SV from the Google Matrix estimates symbol probabilities by padding some probabilities to those low- or zero- probable cells of the stochastic matrix.

Last but not least, how to choose an appropriate data discretization technique may be beyond the discussion of this paper, but we can reasonably conclude that the data discretization methods used in symbolizing data also has a great impact on the performance of these measures. In Section 4.2 and 4.3, we presented the results by using SAX, the combination of PCA and SAX, and simple cut-points as the examples. It seems that finer data granularity (higher $k$) would result in lower AUCs for all the measures (except $D_{GJS}$). This is not always the case. Better discretization rules identified by the domain experts or pattern classification methods could help discover the lurking sequential pattern dynamics, which may also result in higher AUCs regardless of the number of possible symbols ($k$).

The process monitoring from the point of view of sequence divergence/deviation is unique but of similar recent challenges and limitations of discrete sequence anomaly detections [47]. Here, we summarize the limitations and discuss the common challenges of proposed approach.

- The proposed approach may perform relatively poorly when it is applied to sequence data with a few transitions, as the approach is based on evaluating the stability of the system we monitor.
- A fixed length monitoring window ($\Delta[i,j]$) is used to identify anomalies here. However, the way to label anomalies still varies on different applications. And the anomaly labels in sequence data are hard to obtained and might be erroneous [47], especially those identified by humans.
- The length of total discrete temporal sequences ($N$) used to compute the generalized Jensen-Shannon Divergence $D_{GJS}$ and to estimate its significant threshold $D_{GJS|k,m}$ must be large enough to avoid computing the Chi-square statistic in Eq. 15 that may commit a Type II error.
- How to define optimal length of each sequence still rely on domain experts [47]. And we suggested using equally-sized sequences (equally-weighted probability distributions) to obtain optimal estimate of $D_{GJS}$, which might limit the applications of the proposed approach.
- How to choose an appropriate data reduction/discretization technique remain a challenging and open question. This topic has drawn attentions of data mining communities and deserves further research.

## 5 Conclusions

We proposed a novel approach of process monitoring by monitoring the dynamics of a symbolic data stream, which are the patterns of interests identified by the domain experts, pattern classification methods, or time-series representation techniques. We begin with a brief introduction of data discretization. Four important properties of a measure used in monitoring systems—the Boundedness, Symmetry, Generalizability, and Weightability, are also discussed. We suggest using the Steady-State Vectors (SVs) to estimate the discrete system state probability distributions in different times. The generalized Jensen-Shannon Divergence ($D_{GJS}$) is used to assess the differences among discrete temporal sequences by comparing the symbols probability distributions from the SVs. We demonstrate that the $D_{GJS}$ is an outstanding measure to monitor system dynamics and assess the significance of deviation in probabilistic manner. The combination of $D_{GJS}$ and SV as the measure in the monitoring system is proved to outperform others.

### Acknowledgements

We would like to thank editor and anonymous reviewers for their time and feedback that help improve this paper. We would also like to thank Dr. Tatjana Welzer in the Institute of Informatics, University of Maribor, for her advice on data integration and analysis.

## Author Biographies

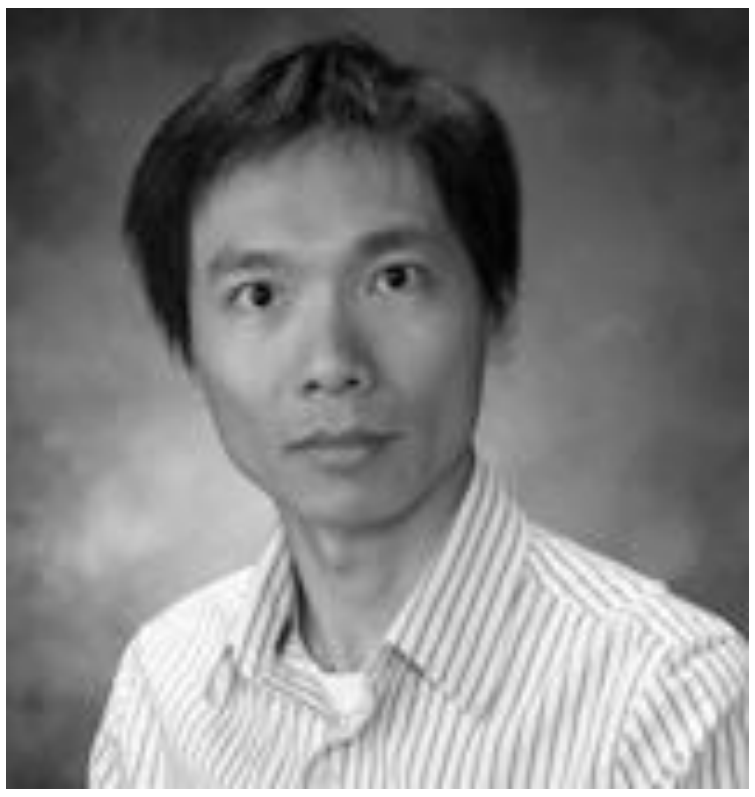

**Yihuang Kang** is an assistant professor in the Department of Information Management, National Sun Yat-sen University. He received his M.S. and Ph.D. in Information Sciences from the iSchool of the University of Pittsburgh in 2007 and 2014. In 2009, He began his work as a data analyst in the University of Pittsburgh School of Medicine. He joined National Sun Yat-sen University in February 2015. His research interests include temporal data mining, business process mining, complex adaptive information system, and healthcare data analytics.

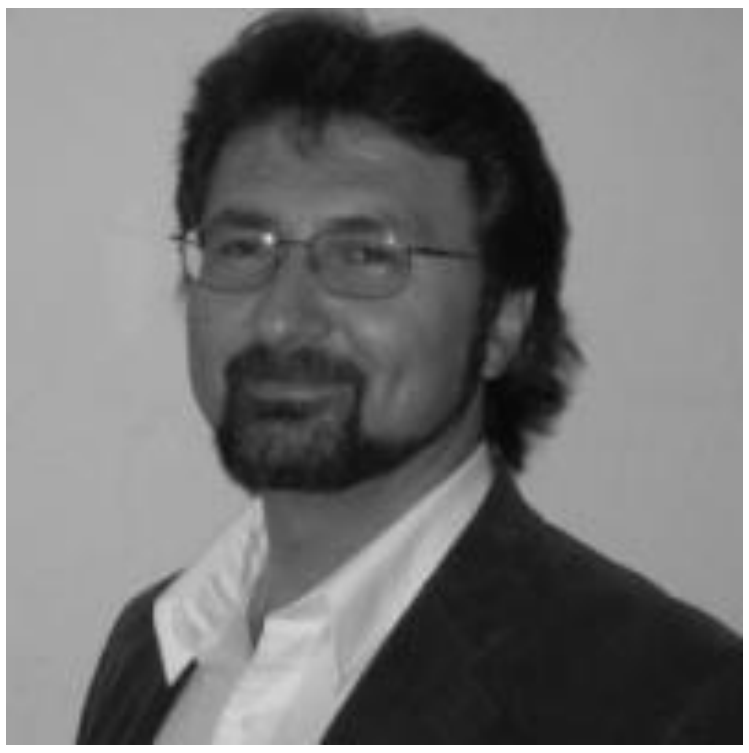

**Vladimir I. Zadorozhny** is an associate professor in the School of Information Science, University of Pittsburgh. He received his Ph.D. in Computer Science in 1993 from the Institute for Problems of Informatics, Russian Academy of Sciences in Moscow. Before coming to the USA, he was a Principal Researcher in the Institute of System Programming, Russian Academy of Sciences. In May 1998, he began his work as a Research Associate in the University of Maryland Institute for Advanced Computer Studies at College Park. He joined the University of Pittsburgh in September 2001. His research interests include complex adaptive information systems, data-intensive process monitoring, information fusion, and scalable architectures for heterogeneous networked information systems.